\documentclass{article}

\usepackage{microtype}
\usepackage{graphicx}
\usepackage{subcaption}
\usepackage{booktabs} % for professional tables
\newcommand{\method}{\text{CraEG}}
\usepackage{hyperref}

\usepackage[preprint]{icml2026}

\usepackage{amsmath}
\usepackage{amssymb}
\usepackage{mathtools}
\usepackage{amsthm}

\usepackage[capitalize,noabbrev]{cleveref}

\theoremstyle{plain}
\newtheorem{theorem}{Theorem}[section]

\theoremstyle{definition}
\newtheorem{definition}[theorem]{Definition}

\theoremstyle{remark}

\usepackage[textsize=tiny]{todonotes}

\icmltitlerunning{Submission and Formatting Instructions for ICML 2026}

\usepackage{algorithm}
\usepackage{algorithmic}
\usepackage{booktabs}
\usepackage{amsmath}
\usepackage{arydshln} 
\usepackage{times}
\usepackage{latexsym}

\usepackage{booktabs}  % 用于美观的表格
\usepackage{multirow}  % 允许单元格跨行
\usepackage{siunitx}   % 用于数字对齐
\usepackage{graphicx}  % 允许表格自动调整宽度
\usepackage{xtab}      % 允许表格跨页
\usepackage{longtable}
\usepackage{color}
\usepackage{colortbl}
\usepackage{xcolor}

\begin{document}

\twocolumn[
  \icmltitle{Decoding in Geometry: Alleviating Embedding-Space Crowding for \\ Complex Reasoning}



  \icmlsetsymbol{equal}{*}

  \begin{icmlauthorlist}
    \icmlauthor{Yixin Yang}{xxx}
    \icmlauthor{Qingxiu Dong}{xxx}
    \icmlauthor{Zhifang Sui}{xxx}
  \end{icmlauthorlist}

  \icmlaffiliation{xxx}{State Key Laboratory of Multimedia Information Processing, School of Computer Science, Peking University}

  \icmlcorrespondingauthor{Yixin Yang}{yangyx@stu.pku.edu.cn}
  \icmlcorrespondingauthor{Zhifang Sui}{szf@pku.edu.cn}

  \icmlkeywords{Machine Learning, ICML}

  \vskip 0.3in
]



\printAffiliationsAndNotice{}  % no special notice (required even if empty)

\begin{abstract}
  % Large language models (LLMs) rely on sampling-based decoding for open-ended generation and reasoning, where decoding strategies critically shape model behavior. 
  % Large language models (LLMs) rely on sampling-based decoding for reasoning, where decoding strategies critically shape model behavior.
  Sampling-based decoding underlies complex reasoning in large language models (LLMs), where decoding strategies critically shape model behavior.
  % Temperature- and truncation-based sampling methods reshape the next-token distribution through global probability reweighting or truncation to balance the quality-diversity tradeoff.
  Temperature- and truncation-based methods reshape the next-token distribution through global probability reweighting or thresholding to balance the quality-diversity tradeoff.
  % However, they treat tokens as independent probability masses and ignore their structural relationships in embedding space.
  % However, they operate solely on token probabilities, ignoring token relationships in embedding space.
  However, they operate solely on token probabilities, ignoring fine-grained relationships among tokens in the embedding space.
  We uncover a novel phenomenon, \textit{embedding-space crowding}, where the next-token distribution concentrates its probability mass on geometrically close tokens in the embedding space.
  %We quantify crowding at multiple granularities and find a statistical association between crowding and reasoning success in mathematical problem solving.
  We quantify crowding at multiple granularities and find a statistical association with reasoning success in mathematical problem solving.
  Motivated by this finding, we propose \method{}, a plug-in sampling method that mitigates crowding through geometry-guided reweighting.
  % Motivated by this finding, we propose \method{}, a plug-in sampling method for mitigating crowding through geometry-guided reweighting.
  % \method{} is training-free and compatible with standard sampling strategies.
  \method{} is training-free, single-pass, and compatible with standard sampling strategies.
  Experiments on multiple models and benchmarks demonstrate improved generation performance, with gains in robustness and diversity metrics.

\end{abstract}

\section{Introduction}
Large language models (LLMs) have recently made substantial progress in complex reasoning~\cite{kaplan2020scaling,achiam2023gpt}.
In these settings, model outputs are largely determined by autoregressive decoding, where token sampling serves as the fundamental operation. 
Decoding strategies can therefore have a significant impact on the resulting generations~\cite{shi2024thorough}. 
Sampling-based decoding has proven particularly effective for LLMs by selecting each next token from the model's predicted next-token distribution.  
It remains the prevailing method for open-ended generation and reasoning~\cite{chowdhery2023palm,guo2025deepseek}.
Prior work mainly falls into two categories: truncation-based sampling (e.g., top-$p$~\cite{holtzman2019curious}, top-$k$~\cite{fan2018hierarchical}) and temperature-based sampling (e.g., % AdapT~\cite{zhu2024hot}, 
Temperature Scaling~\cite{ackley1985learning}, 
EDT~\cite{zhang2024edt}). 
By modulating the cutoff threshold or the temperature parameter, these methods globally modify the token probability distribution through truncation, sharpening, or smoothing to better balance the \textit{quality-diversity tradeoff}.
In this work, we study decoding through a geometry-aware view of the next-token distribution, and diagnose its effect on generation behavior.
% In particular, we uncover a novel and underexplored phenomenon, which we term \textit{embedding-space crowding}. 
We uncover a novel and underexplored phenomenon, which we term \textit{embedding-space crowding}. 
This phenomenon occurs when the next-token distribution concentrates its probability mass in a narrow region of the embedding space.
% We formalize and quantify embedding-space crowding, and analyze its relationship with reasoning outcomes. 
We formalize and quantify embedding-space crowding at multiple granularities, and analyze its relationship with reasoning outcomes. 
Specifically, we use mathematical reasoning as a controlled testbed, where the task is well-defined and correctness is automatically verifiable.  
% Across AIME problems with Qwen, our analysis shows that both sequence-level and step-level crowding are statistically associated with final answer correctness.
Across the AIME benchmark with the Qwen model, our analysis shows that both sequence-level and step-level crowding score are statistically associated with final answer correctness.
% These findings suggest that embedding-space crowding is significantly associated with reasoning success, while remaining largely unaddressed in prior decoding approaches.
These findings suggest a statistically significant association between embedding-space crowding and reasoning success, a factor largely unaddressed by prior decoding approaches.

Inspired by this discovery, we propose \textbf{Cr}owding-\textbf{A}ware Sampling via \textbf{E}mbedding \textbf{G}eometry (\method{}), a novel plug-in decoding method that alleviating embedding-space crowding and improves the quality-diversity tradeoff in generation. 
% CraEG performs geometry-guided, crowding-aware reweighting with step-adaptive correction strength.
% Specifically, \method{} constructs a geometry-guided reweighting scheme  with step-adaptive correction strength for the next-token distribution, downweighting tokens that simultaneously exhibit high probability and high crowding. 
% Specifically, \method{} applies a geometry-guided reweighting scheme with step-adaptive correction strength to the next-token distribution, downweighting tokens that simultaneously exhibit high probability and high crowding.
Specifically, \method{} applies a geometry-guided reweighting scheme to the next-token distribution, downweighting higher-probability, higher-crowding tokens with step-adaptive strength.
%By reducing redundancy in the embedding space, \method{} promotes sampling beyond locally crowded region, enabling more robust and diverse reasoning trajectories.
%This reduction in redundancy promotes sampling beyond narrow, crowded regions, enabling more diverse and robust reasoning trajectories.
By reducing redundancy in the embedding space, \method{} promotes sampling beyond narrow, crowded regions, enabling more diverse and robust reasoning trajectories.
%\method{} is training-free and can be seamlessly combined with standard sampling strategies such as top-$p$, top-$k$, and temperature scaling.
% Moreover, \method{} is training-free and is compatible with standard sampling strategies such as top-$p$ and temperature scaling.
% Moreover, \method{} is training-free, requires no additional supervision, and introduces no extra forward passes. 
Moreover, \method{} is training-free, requires no additional supervision or auxiliary models, and introduces no extra forward passes. 
It is also compatible with common sampling strategies such as top-$p$ and temperature scaling.

%We demonstrate the effectiveness of \method{} across multiple models and evaluation settings.
%Experiments are conducted with Qwen3-1.7B, Qwen3-4B, and Hunyuan-1.8B on three mathematical reasoning benchmarks.
% We consider standard sampling configurations, using top-$p$ sampling and temperature scaling as strong baselines. 
%We adopt standard sampling configurations, with top-$p$ sampling and temperature scaling as strong baselines.
%Overall, \method{} improves generation performance across models and benchmarks, leading to improvements in robustness and diversity metrics in many settings.
We demonstrate the effectiveness of \method{} across multiple models and evaluation settings. 
Experiments are conducted with Qwen3-1.7B~\cite{yang2025qwen3}, Qwen3-4B, and Hunyuan-1.8B~\cite{tencent_hunyuan_18b_instruct} on three challenging mathematical reasoning benchmarks. 
We adopt standard sampling configurations, with top-$p$ sampling and temperature scaling as strong baselines.
We find that \method{} improves generation performance across diverse settings, with gains on robustness and diversity metrics.
% For instance, on Qwen3-1.7B with temperature $=1$ and top-$p=1$, \method{} improves avg@32 by 0.52 points and pass@8 by 1.98 percentage points, with gains of 1.17 points in distinct-$n$ and 0.62 points in semantic diversity.
For instance, on Qwen3-1.7B (temperature $=1$, top-$p=1$), \method{} improves avg@32 by 0.52 points and pass@8 by 1.98 percentage points, with gains of 1.17 points in distinct-$n$ and 0.62 points in semantic diversity.
Our main contributions are as follows:
%The main contributions of our paper are as follows:
\begin{itemize}
    \item We uncover and formalize \textit{embedding-space crowding}, a decoding phenomenon where next-token probability mass concentrates locally in embedding space.
    %We uncover and formalize \textit{embedding-space crowding}, a novel decoding phenomenon where the next-token probability mass concentrates within a local region of the embedding space.
    \item We propose quantitative crowding measures and show their statistical association with reasoning success on challenging mathematical reasoning benchmarks.
    % We propose quantitative measures of crowding and show that crowding is statistically associated with reasoning success on challenging mathematical reasoning benchmarks.
    \item We introduce \method{}, a training-free, plug-in decoding method that uses geometry-guided reweighting to alleviate crowding and reduce embedding redundancy.
    % We introduce \method{}, a training-free, plug-in decoding method that performs geometry-guided reweighting to mitigate crowding and reduce embedding redundancy.
    \item We demonstrate the effectiveness of \method{} across multiple models, benchmarks and evaluation settings, improving robustness and diversity metrics under standard sampling baselines.
    % We demonstrate the effectiveness of \method{} across multiple models and evaluation settings, achieving improved robustness and diversity metrics under standard sampling configurations.
\end{itemize}

% 第三段
%In this work, we adopt a geometric perspective on decoding. % 不同的视角
%We hypothesize that embedding-space crowding reflects a decoding regime with reduced effective exploration capacity, where multiple high-probability tokens are geometrically correlated and collectively constrain the next-token choice. % 猜想
%While geometric concentration does not deterministically imply incorrectness, we find that decoding states with higher embedding-space crowding are statistically more likely to result in incorrect reasoning outcomes. % 防御
% 我们做了什么内容（一分析、二decoding修正）

% 第四段
% 实验结果——分析的结果、修正实验结果

%Our contributions are threefold:
%\begin{itemize}
%  \item we define and analyze embedding-space crowding as a novel geometric property of the next-token distribution;
%  \item we empirically demonstrate a strong association between crowding and reasoning failure across math reasoning benchmarks;
%  \item we propose a crowd-aware decoding correction that mitigates geometric over-concentration and improves both accuracy and diversity.
%\end{itemize}

\begin{figure}[ht]
  \begin{center}
    \centerline{\includegraphics[width=\columnwidth]{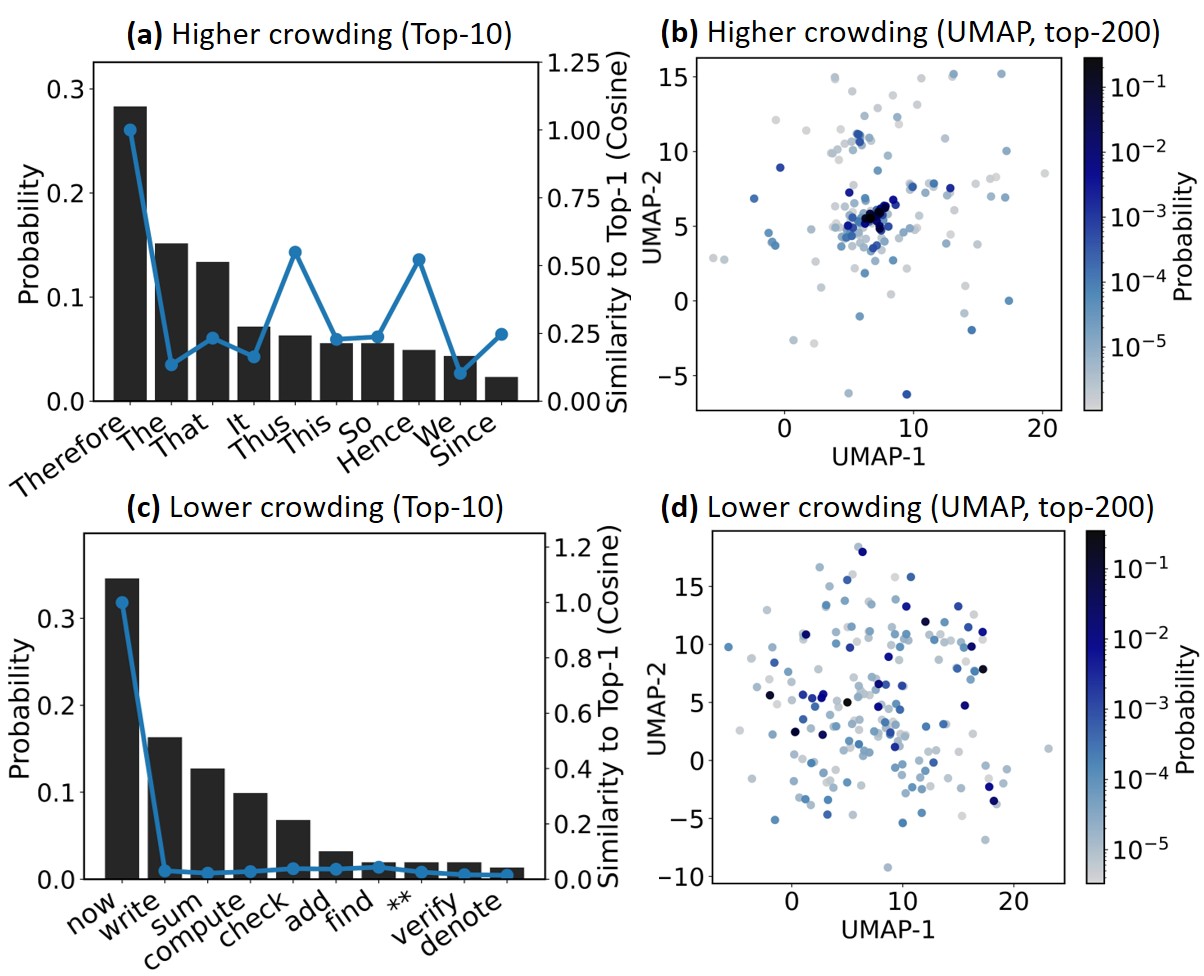}}
    \caption{Examples of embedding-space crowding in next-token distributions. We compare two decoding states from the same AIME25 solution trajectory with similar entropy but different crowding levels (higher: Step 447; lower: Step 2003). Left panels show the top-10 probabilities and similarity to the top candidate; right panels show a UMAP projection of the top-200 tokens colored by probability. Higher crowding concentrates probability mass on geometrically similar tokens; lower crowding is more dispersed.
    }
    \label{fig:crowding_example}
  \end{center}
\end{figure}

\section{Related Work}
\paragraph{Sampling-Based Decoding Methods.}
% 简洁化
% Sampling-based decoding is the dominant paradigm for open-ended text generation 和推理任务 with large language models. Sampling-based decoding为了更好的quality-diversity tradeoff，其方法主要包括两类方法。第一类方法truncation-based methods通过限制采样范围，从而避免低概率 token 带来的不稳定性。top-p restricts the token pool to those whose cumulative probability exceeds a total threshold p。top-k和$ε$-sampling分别使用确定的token个数或概率threshold进行筛选。More recent  引入adaptive truncation rules based on uncertainty，使thresholds根据当前分布调整。第二类方法temperature-based采样直接调整 logits 或概率分布的整体形状，从而平衡生成的质量和多样性。Temperature scaling 通过调整温度参数大小 adjusts distribution 整体 sharpness。recent 一些研究 动态调整每个decoding step上的温度based on uncertainty signal，如entropy, confidence, or perplexity。（然而这些方法……（说问题））除了上述外，还有一些其他的Sampling-Based Decoding Methods。Guided Decoding  结合外部引导信号来引导生成所需的属性。Contrastive Decoding通过使用额外模型或额外 forward pass提供对比从而提升质量。这些方法虽然some overlapping aims as our work，它们需要外部或辅助的偏好信号，be viewed as 与\method{}和truncation-based、temperature-based方法中可以结合使用的互补方法。
% 外部或辅助的偏好信号
% Top-$p$~\cite{} selects tokens with cumulative probability above $p$, while top-$k$ and $\epsilon$-sampling filter by a fixed token count or probability threshold. 
Sampling-based decoding dominates open-ended generation and complex reasoning in LLMs~\cite{guo2025deepseek,shi2024thorough,chowdhery2023palm}. To achieve a better quality-diversity tradeoff, existing methods mainly fall into two categories. Truncation-based methods limit the sampling space to avoid instability from low-probability tokens. Top-$p$~\cite{holtzman2019curious} selects tokens with cumulative probability above $p$, while top-$k$~\cite{fan2018hierarchical} and $\epsilon$-sampling~\cite{freitag2023epsilon} use fixed counts or probability thresholds. Recent methods further adapt truncation thresholds based on different uncertainty signals, such as entropy~\cite{hewitt2022truncation, tan2025p}, confidence~\cite{nguyen2024turning}, or perplexity~\cite{basu2020mirostat}. Temperature-based methods modify the overall probability distribution by scaling logits or probabilities with the temperature parameter. Temperature scaling~\cite{ackley1985learning} applies a fixed temperature. Other works~\cite{zhang2024edt, zhu2024hot} dynamically adjusts the temperature at each decoding step based on uncertainty signals. Recent studies~\cite{wang2025end,troshin2025control} further employ lightweight models to dynamically predict suitable temperature and truncation thresholds during decoding. However, such methods remain restricted to coarse-grained distribution control, ignoring the properties of individual tokens. Other approaches, including guided~\cite{dathathri2019plug,pynadath2025controlled} and contrastive decoding~\cite{liu2021dexperts,zhao2024enhancing}, use external signals or auxiliary models to control attributes or enhance generation quality. These methods differ substantially by requiring external or auxiliary preference signals and are complementary to \method{}, as well as to truncation- and temperature-based approaches.

\paragraph{Geometry of Token Representations in LLMs.}
% embedding geometry 被研究过;但多在 training / representation 层;很少触及 decoding-time distribution structure
% Token embeddings form a fundamental backbone of LLMs. 
% Token embeddings form the fundamental backbone of LLMs, and the geometry of token representations has been extensively studied. Some works explore the intrinsic geometry of token embedding spaces in LLMs, highlighting properties such as dimensionality~\cite{}, anisotropy~\cite{}, stratified manifolds~\cite{}, and hierarchical structures~\cite{}. Others focus on the geometric alignment and transferability of token embeddings across models of different scales, modalities~\cite{}, and languages~\cite{}. Additionally, studies~\cite{} examine anomalies in token representations and their impact on embedding geometry and model performance. Some works~\cite{} investigate how the geometry and topology of token embeddings affect downstream task outcomes. Prior work on token embedding geometry has focused primarily on static representations. In contrast, we study how embedding geometry interacts with probability mass at decoding time to shape generation trajectories.
Token embeddings form the backbone of LLMs, and their geometry has been extensively studied. Some works explore the intrinsic geometry of token embedding spaces, highlighting properties such as dimensionality~\cite{kataiwa2025measuring}, anisotropy~\cite{godey2024anisotropy}, stratified manifolds~\cite{robinson2024structure}, and hierarchical structures~\cite{park2024geometry}. Others examine the geometric alignment and transferability of token embeddings across models of different scales, modalities~\cite{wang2024llms}, and languages~\cite{lee2025shared}. Additionally, studies~\cite{chen2025sticking,robinson2025token} investigate anomalies in token representations and their impact on embedding geometry and model performance. Some works~\cite{zhao2024implicit,robinson2025probing} explore how token embedding geometry and topology influence downstream tasks. Prior work has focused primarily on static representations, while we investigate how embedding geometry interacts with probability mass during decoding to shape generation trajectories.

% explore the intrinsic geometry of token embedding spaces in LLMs, highlighting properties such as dimensionality, anisotropy, stratified manifolds, and hierarchical structures.

% Prior work on token embedding geometry has focused primarily on static representations or internal dynamics during training. In contrast, we study how embedding geometry interacts with probability mass at decoding time to shape generation trajectories. 
% Token embeddings form a fundamental backbone of LLMs. Geometry of token representation 已经有许多研究。一些研究explore the intrinsic geometry of token embedding spaces in LLMs, highlighting properties such as dimensionality, anisotropy, stratified manifolds, and hierarchical structures。另一些研究关注token embedding 在不同模型间的几何对齐、迁移性和共享结构，探讨了不同模型，如跨规模、跨模态、跨语言，之间的表示是否可以共享或迁移。还有研究聚焦于token 表示中存在的异常现象，研究它们如何破坏嵌入空间的几何结构和模型的表现。一些文章分析了 token 表示的几何结构与下游任务之间的关系，探讨了嵌入空间的几何与拓扑特性如何影响任务表现。

\paragraph{Inference-Time Methods for Reasoning in LLMs.}
% 长推理任务中的推理增强方法通常依赖于在推理过程中进行额外干预，以提升模型的推理质量和多样性。Self-consistency 和 Best-of-N，通过生成多个候选并根据一致性或覆盖率选择最优结果；前瞻采样（Lookahead Sampling） 和 搜索方法，通过推理时模拟未来步骤的效果来优化当前决策；以及 工具调用 和 外部验证器，它们通过引入外部信号对模型输出进行修正或后处理。这些方法有效地增强了模型的推理能力，但它们通常需要额外的计算资源（如多次推理、外部模型或工具的支持），使得计算开销和复杂度大大增加。与这些方法不同，本文的工作聚焦于解码阶段的单步重加权，无需外部信号或额外模型计算。我们提出的 crowd-aware 解码策略通过分析 next-token 分布的几何结构，直接调整模型内部分布，从而提升长推理任务的稳健性与准确性，同时保持较低的计算开销。
% 在推理过程中，许多方法通过对生成过程进行干预来提高模型的reasoning质量和多样性。
% 这些方法有效地增强了模型的推理能力，但它们通常需要额外的计算资源（如多次推理、外部模型或工具的支持），使得计算开销和复杂度大大增加。
% 这些方法通常能够有效提升推理的稳定性和多样性，但需要额外的计算资源和复杂度。
% 许多方法通过对生成过程进行干预来提高模型的推理质量和多样性。self-consistency 和 Best-of-N，它们通过生成多个候选并根据一致性或覆盖率选择最优结果。前瞻采样（Predictive Sampling） 和 搜索方法（Search-Based Reasoning Methods），它们通过推理时模拟未来步骤的效果来优化当前决策。Guided Decoding 方法通过在每一步进行自评估（self-evaluation）来引导解码过程。以及 工具调用 和 外部验证器，通过引入外部信号对模型输出进行修正或后处理。While these methods enhance reasoning capabilities, 它需要额外的计算资源，如多次推理、外部模型或工具的支持。这使它们在复杂度上与我们和truncation- and temperature-based sampling methods本质上区分开来。与这些方法不同，本文的工作 提出crowd-aware 解码策略，聚焦于解码阶段的单步重加权，无需外部信号或额外模型计算。
Many methods improve reasoning quality and diversity by intervening in the generation process. Self-consistency~\cite{wang2022self} and Best-of-N~\cite{shi2025semantic,meyerson2025solving} generate multiple candidates and select the optimal result based on consistency or coverage. Predictive Sampling~\cite{ma2024non,xu2025phi} and Search-Based Reasoning~\cite{yao2023tree,hamm2025novelty} optimize current decisions by simulating future steps. Self-Refine~\cite{madaan2023self} and Guided Decoding~\cite{xie2023self} use self-evaluation to iteratively improve output or guide decoding. Additionally, approaches like tool calls~\cite{yao2022react} and external validators~\cite{ling2023deductive} adjust or post-process model outputs using external signals. These methods enhance reasoning but differ significantly in complexity, making them complementary to approaches like \method{}, truncation, and temperature scaling. In comparison, our work introduces a crowd-aware decoding strategy, focusing on single-step reweighting without external signals or additional model computations.

% While enhancing reasoning, they differ substantially in their level of complexity and can be viewed as complementary approachs. In contrast, our work introduces a crowd-aware decoding strategy, focusing on single-step reweighting without external signals or additional model computations.

% \paragraph{Decoding Behavior and Reasoning Failures in LLMs}
% diversity / mode collapse 是 outcome; 你关注的是 mechanism

% Our work complements studies on diversity and mode collapse by identifying a decoding-time geometric mechanism that constrains effective exploration at the token level.

\section{Embedding-space Crowding in Decoding}
\label{sec:define_analysis}

In this section, we identify an underexplored phenomenon in LLM decoding, which we term embedding-space crowding.
% Embedding-space crowding characterizes decoding states in which 
% probability mass is concentrated on tokens that are geometrically close in embedding space. 
Embedding-space crowding characterizes decoding states in which probability mass is concentrated on tokens that are geometrically close in embedding space, as illustrated in Figure~\ref{fig:crowding_example}.
% (Figure~\ref{fig:crowding_example}).
% We formally define this notion and introduce a quantitative measure to characterize the degree of crowding. Empirically, we find that higher embedding-space crowding is consistently associated with a higher likelihood of reasoning failure.
% Notably, this geometric concentration is invisible to probability-only decoding controls, including temperature and truncation.
We formally define this notion and introduce quantitative measures of crowding at multiple granularities.
% We formally define this notion and introduce quantitative measures to characterize the degree of crowding. 
% Empirically, we find that higher embedding-space crowding is statistically associated with an increased likelihood of reasoning failure.
Empirically, we find that higher embedding-space crowding is statistically associated with lower reasoning success.

\subsection{Defining Embedding-Space Crowding}
\label{sec:31}
% We define embedding-space crowding as a property of the next-token distribution during decoding, reflecting how probability mass is organized among tokens in embedding space.
%We define embedding-space crowding on the next-token distribution at each decoding step.
%It describes how probability mass is arranged in embedding space.
%Intuitively, crowding is high when a large fraction of probability mass is assigned to tokens whose embeddings are geometrically close to one another, indicating a concentrated distribution over a narrow region of representation space.
%Crowding is high when much of the probability mass falls on tokens that are close in embedding space.
%The distribution concentrates on a small region rather than spreading across distant tokens.
%Unlike uncertainty-based measures that consider only the dispersion of probability mass, crowding explicitly accounts for the geometric relationships between token embeddings.% 确认下uncertainty or dispersion，要伴随调研
We define embedding-space crowding on the next-token distribution at each decoding step. It describes how probability mass is arranged in embedding space. Crowding is high when a large share of mass falls on tokens that are close under a given similarity metric, forming a compact region. 
%At the level of distributional metrics, crowding differs from common uncertainty-based measures by incorporating the geometric relationships between token embeddings. 
Crowding differs from widely-used uncertainty metrics in decoding by accounting for geometric relationships among token embeddings.
In what follows, we formalize this intuition by measuring how probability mass concentrates around geometrically similar tokens.
%measuring how much probability mass lies within local neighborhoods in embedding space.

\begin{definition}[Token-Level Crowding Score]
\label{def:token} 
Consider a language model at a given decoding step $t$ with next-token probability distribution $\{p_{t,j}\}_{j \in V}$ and corresponding token embeddings $\{e_j\}_{j \in V}$. The \emph{token-level embedding-space crowding score} of a token $i$ is defined as
\begin{equation}
\mathrm{Crowd}^{\text{token}}_{t}(i)
\;=\;
\sum_{j \neq i} p_{t,j} \, \lvert \cos(e_i, e_j) \rvert,
\end{equation}
where $V$ denotes the vocabulary and $\cos(\cdot,\cdot)$ denotes cosine similarity between embeddings.
\end{definition}

 %Under Definition~\ref{def:token}, the crowding of a token reflects the probability-weighted geometric density of alternative tokens in its embedding neighborhood. A token has high crowding when it is surrounded by one or more other tokens that are geometrically correlated in embedding space and assigned non-negligible probability mass. Conversely, a token exhibits low crowding when few alternative tokens lie in its geometric vicinity, regardless of whether the overall distribution is peaked or dispersed. 
Under Definition~\ref{def:token}, a token-level crowding score reflects the probability-weighted concentration of alternative tokens relative to token $i$ in embedding space.
%A token has high crowding score when it is accompanied by other geometrically similar tokens that receive non-negligible probability mass. A token exhibits low crowding score when few plausible alternatives are geometrically close. 
A token has high crowding when other geometrically similar tokens carry non-negligible probability mass. It has low crowding when few plausible alternatives lie nearby.
Specifically, we use absolute cosine similarity to measure geometric association strength independent of sign. This treats embeddings pointing in the same or opposite directions as equally associated, since both indicate near-collinearity in embedding space.

\begin{definition}[Step-Level Crowding Score]
\label{def:step} 
Given the token-level crowding scores, we define the \emph{step-level embedding-space crowding score} at decoding step $t$ as
\begin{equation}
\mathrm{Crowd}^{\text{step}}(t)
\;=\;
\sum_i p_{t,i} \cdot \mathrm{Crowd}^{\text{token}}_{t}(i),
\end{equation}
% where $p_{t,i}$ denotes the model-assigned probability of token $i$ at decoding step $t$.
% where $p_{t,i}$ is the model's probability of token $i$ at step $t$.
\end{definition}

% Step-level crowding aggregates token-level crowding under the next-token distribution, and therefore reflects the overall degree of geometric concentration exhibited at a decoding step, as formalized in Definition~\ref{def:step}. High step-level crowding arises when probability mass is distributed among multiple tokens that are geometrically correlated, indicating the presence of meaningful geometric competition among alternative tokens. In contrast, step-level crowding remains low when probability mass is either spread across geometrically diverse regions or dominated by a single token, both of which reflect the absence of substantial geometric redundancy among alternatives. Furthermore, we define the sample-level crowding score by averaging the crowding values across decoding steps, as formalized in Definition~\ref{def:sample}. Sample-level crowding thus characterizes the typical degree of embedding-space crowding encountered during a generation.
Reasoning success is measured at the level of the entire generation sequence. To enable a more direct diagnosis, we further define step-level and sequence-level crowding scores based on the token-level definition.
Step-level crowding score is the expected token-level crowding under the next-token distribution at step $t$ (Definition~\ref{def:step}). It is high when probability mass concentrates on multiple geometrically similar tokens, reflecting competition among close alternatives. It is low when mass either spreads across geometrically diverse tokens or collapses onto a single token, in both cases exhibiting little geometric redundancy.
% It is high when substantial probability mass lies on multiple geometrically similar tokens, indicating strong competition among close alternatives. It is low when mass is spread over geometrically diverse tokens or collapses onto a single token. In either case, there is little geometric redundancy among alternatives. 
% We extend this to the sequence level by averaging step-level crowding across decoding steps (Definition~\ref{def:seq}), yielding the average crowding over an entire generation.
We further extend this to the sequence level by averaging step-level crowding across decoding steps, which yields the average crowding over an entire generation (Definition~\ref{def:seq}). 
%The sequence-level crowding score is an aggregate measure of step-level crowding throughout the entire generation. Specifically, a higher score corresponds to a greater overall degree of crowding, and a lower score corresponds to less.
% A higher sequence-level crowding score corresponds to a greater overall degree of crowding, and a lower score corresponds to less.

\begin{figure}[ht]
  \begin{center}
    \centerline{\includegraphics[width=\columnwidth]{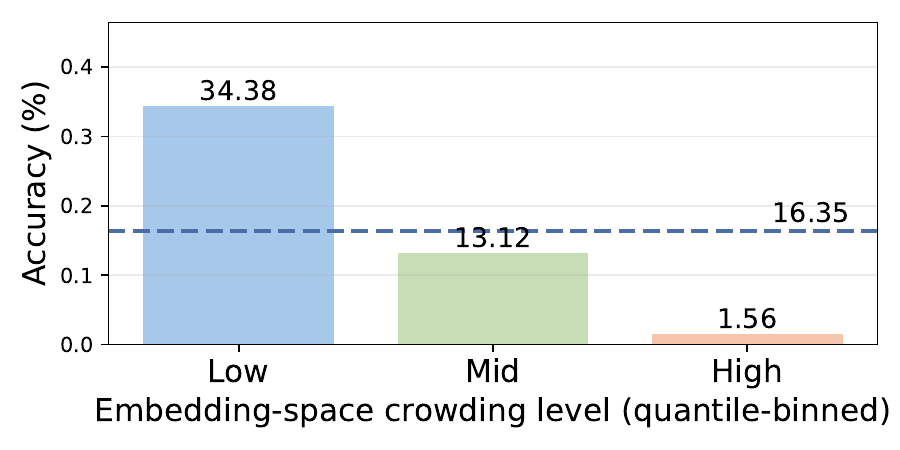}}
    \caption{%Sequence-level embedding-space crowding and reasoning accuracy.
%Samples are partitioned into low, mid, and high crowding groups using quantile-based binning. Bars report accuracy within each group. Accuracy decreases monotonically with increasing crowding, indicating a strong negative association between sequence-level crowding and reasoning correctness.Sequence-level embedding-space crowding and reasoning accuracy.
% Accuracy decreases monotonically with crowding, consistent with a negative association between sequence-level crowding and correctness. 
Sequences are binned into low/mid/high groups by quantiles of sequence-level crowding. Bars report accuracy per bin, which decreases monotonically with crowding.
%Based on the sequence-level crowding score, sequences are partitioned into low, mid, and high crowding groups through quantile-based binning. Bars show accuracy in each group. Accuracy decreases monotonically with the sequence-level crowding score, demonstrating an inverse relationship.
}
    \label{fig:3bin}
  \end{center}
\end{figure}

\begin{definition}[Sequence-Level Crowding Score]
\label{def:seq}
For a generated sequence consisting of $T$ decoding steps, we define the \emph{sequence-level embedding-space crowding score} as
\begin{equation}
\mathrm{Crowd}^{\text{seq}}
\;=\;
\frac{1}{T} \sum_{t=1}^T \mathrm{Crowd}^{\text{step}}(t),
\end{equation}
%where $T$ denotes the total number of decoding steps in the generated sequence.
\end{definition}

% In practice, computing crowding over the full vocabulary is unnecessary, as tokens with extremely low probability contribute negligibly to the crowding score. We therefore approximate crowding by restricting computation to a candidate subset $S\subset V$, consisting of the top-$K$ tokens by probability at each decoding step. This approximation substantially reduces computational cost while preserving the geometric structure of high-probability regions in the embedding space. In our experiments, we use a fixed candidate size $K$ ($k=100$), which we find sufficient to capture the geometric structure of high-probability regions.

% In practice, computing step-level crowding over the full vocabulary is unnecessary, as tokens with extremely low probability contribute negligibly. We therefore approximate crowding by restricting computation to a token subset $S\subset V$, consisting of the top-$K$ tokens by probability at each decoding step. This approximation substantially reduces computational cost while focusing computation on regions of the next-token distribution with non-negligible probability mass. In all analysis experiments, we use a fixed value of $K$.
In practice, computing crowding scores over the full vocabulary is unnecessary, since low-probability tokens contribute negligibly.
% We approximate crowding by restricting computation to a token subset
We approximate the sums in our crowding scores by restricting computation to a subset $S\subset V$ that contains the top-$K$ tokens under the next-token distribution at each step. This greatly reduces computational cost while retaining most of the probability mass. We use a fixed $K$ throughout our analyses in Section~\ref{sec:ea}.

%Unless otherwise specified, we set K=100 in all experiments; we found similar trends for nearby values of K.

\subsection{Empirical Analysis of Embedding-Space Crowding}
\label{sec:ea}

We study embedding-space crowding on mathematical reasoning using the AIME25 benchmark~\cite{aime2025}, which contains 30 problems with exact numeric answers. For each problem, we generate 32 samples with temperature 1.0 and top-$p$ 1.0, yielding 960 reasoning traces. %We use Qwen3-0.6B~\cite{yang2025qwen3} as a computationally tractable case study for token-level analysis.
We use Qwen3-0.6B~\cite{yang2025qwen3} as a case study because it offers basic complex-reasoning capability while remaining computationally tractable.
At each decoding step, we compute token- and step-level crowding from the next-token distribution over the top-$K$ tokens ($K=100$). Sequence-level crowding is the average step-level crowding over the full generation, as defined in Section~\ref{sec:31}. We evaluate correctness by exact matching of the final numeric answer~\cite{hendrycks2021measuring}. Our analysis at both the sequence and step levels confirms a significant negative association between 
%embedding-space
crowding and reasoning success. %We analyze the association between crowding and correctness at both the sequence and step levels. Our analysis above establishes a significant statistical association between embedding-space crowding and reasoning success. 

% \noindent\textbf
\paragraph{Sequence-level analysis.} 
%We examine how sequence-level embedding-space crowding relates to reasoning correctness. We bin samples into three equal-sized quantile groups (low, mid, high) by crowding. Accuracy decreases monotonically with crowding, from 34.37\% (low) to 13.13\% (mid) and 1.56\% (high), as shown in Figure~\ref{}. The overall accuracy is 16.4\%. This monotonic decrease is not an artifact of binning. Treating crowding continuously, correctness is negatively correlated with crowding under point-biserial correlation ($r=-0.392$, $p=1.38\times10^{-36}$). Overall, higher sequence-level crowding is consistently associated with lower correctness.
We analyze the association between sequence-level crowding and reasoning correctness. We bucket sequences into tertiles (low/mid/high) by crowding to form equal-sized groups. Figure~\ref{fig:3bin} shows that accuracy decreases monotonically with crowding, dropping from 34.37\% (low) to 13.13\% (mid) and 1.56\% (high). % This monotonic decrease is not an artifact of binning. Treating crowding continuously, correctness is negatively correlated with crowding under point-biserial correlation ($r=-0.392$, $p=1.38\times10^{-36}$). 
We further validate this trend by treating crowding continuously. The point-biserial correlation between crowding and correctness is significantly negative ($r = -0.39$, $p = 1.38 \times 10^{-36}$).
Overall, these results indicate a significant negative association between embedding-space crowding and reasoning success.% at the sequence level.
% This finding is further validated by our analysis on the AIME24 benchmark~\cite{aime2024} (see Appendix~\ref{app:other_d}).

% We first examine the relationship between sample-level embedding-space crowding and reasoning correctness. To provide an interpretable comparison, we partition samples into three groups (low, mid, and high crowding) using quantile-based binning, such that each group contains an equal number of samples. As shown in Figure~\ref{}, we observe a pronounced monotonic trend across crowding levels. Samples in the low-crowding group achieve substantially higher accuracy (34.37\%) compared to those in the mid-crowding group (13.13\%), while samples with high embedding-space crowding exhibit a near-complete collapse in accuracy (1.56\%). This represents a sharp degradation relative to the overall mean accuracy of 16.4\%, indicating that high crowding is strongly associated with reasoning failure. This trend persists when crowding is treated as a continuous variable. Sample-level crowding is significantly negatively correlated with correctness according to both Pearson ($r=-0.39$) and Spearman ($\rho=-0.40$) correlations, with highly significant p-values (both $p < 10^{-30}$). Together, these results demonstrate a robust empirical association between embedding-space crowding and reasoning outcomes at the sample level.
%进一步，我们还在hunyuan-0.6B模型和AIME24数据集上进行分析实验
% \noindent\textbf

\paragraph{Step-level analysis.} 
To test whether the sequence-level association between crowding and correctness is driven by a small fraction of high-crowding steps, we conduct a complementary step-level analysis. We pool decoding steps across all samples and group them by the final correctness of their parent sample. 
% The step-level ECDFs in Appendix~\ref{fig:ecdf} show a general rightward shift toward higher crowding for steps from incorrect samples. We treat this comparison descriptively. 
As a descriptive comparison, the step-level ECDFs in Appendix~\ref{fig:ecdf} show a general rightward shift toward higher crowding for steps from incorrect samples.
Elevated crowding is broadly present across steps rather than confined to a few extremes.
% To test whether the sequence-level association between crowding and correctness is driven by a small fraction of high-crowding steps, we conduct a complementary step-level analysis. We pool decoding steps and group them by the final correctness of their parent sample. Figure~\ref{fig:ecdf} plots ECDFs of step-level crowding for steps drawn from correct and incorrect samples. Across most quantiles, steps from incorrect samples exhibit higher crowding than those from correct samples. This comparison is descriptive, as decoding steps within a sample are not independent, and indicates that elevated crowding is not confined to a small number of extreme steps.
%To assess whether the sample-level relationship between embedding-space crowding and reasoning outcomes is driven by a small number of extreme decoding steps, we perform a supporting analysis at the step level. We aggregate all decoding steps from the generated samples and compare their crowding values based on the final correctness of the corresponding sample. Figure~\ref{} shows the empirical cumulative distribution functions (ECDFs) of step-level crowding for correct and incorrect samples. Over most of the distribution, decoding steps from incorrect samples tend to have higher crowding values than those from correct samples. This pattern suggests that higher crowding is not limited to a few isolated steps, but is instead observed more broadly throughout the decoding process of incorrect samples.

\paragraph{Relationship with uncertainty.} %We then further examine the relationship between embedding-space crowding and conventional uncertainty measures. We use the Shannon entropy~\cite{shannon1948mathematical} of the next-token distribution as a decoding-time uncertainty measure. Entropy summarizes dispersion of probability mass, whereas crowding score depends on its organization in embedding space. We fit a logistic regression predicting correctness from standardized sequence-level crowding score and sequence average entropy. Crowding score remains a significant negative predictor of correctness (odds ratio $=0.29$, $p=0.001$), while entropy is not significant (odds ratio $=0.63$, $p=0.26$). 
%This indicates that the observed association is not explained by entropy alone. 
%Thus, even after controlling for sequence entropy, crowding score retains a unique and significant association with reasoning success. Embedding-space crowding reflects geometric structure in the distribution that is largely uncaptured by uncertainty measures.
%Full regression results are reported in Appendix~\ref{app:regression}. Appendix~\ref{appendix:vce} visualize the step-level crowding-entropy relationship.
% We then further examine the relationship between embedding-space crowding and conventional uncertainty measures. 
We use the Shannon entropy~\cite{shannon1948mathematical} of the next-token distribution as a decoding-time conventional uncertainty measure. While entropy captures overall dispersion, crowding captures how probability mass is organized in embedding space. A logistic regression with standardized sequence-level crowding and average entropy shows that crowding remains a significant negative predictor of correctness (odds ratio $=0.29$, $p=0.001$), whereas entropy is not significant (odds ratio $=0.63$, $p=0.26$). Full regression results and additional visualizations are provided in Appendix~\ref{app:regression} and Appendix~\ref{appendix:vce}.

\section{Method}
\label{sec:method}
Motivated by the empirical analyses in Section~\ref{sec:31}, we introduce \method{}, a novel plug-in decoding method that alleviates embedding-space crowding and improves the quality-diversity tradeoff in generation. \method{} applies a geometry-aware reweighting scheme to the next-token distribution, downweighting higher-probability, higher-crowding tokens. Given the negative association between crowding and reasoning success, this encourages more robust decoding on complex reasoning tasks. Furthermore, by promoting sampling beyond narrow, crowded regions, \method{} naturally fosters more diverse reasoning trajectories.
% Motivated by the empirical analyses in Section~\ref{sec:31}, we propose a crowd-aware decoding method that explicitly accounts for embedding-space crowding during next-token selection. At each decoding step, the method evaluates how strongly candidate tokens are embedded within locally crowded regions of the token embedding space and penalizes tokens whose probability mass is excessively concentrated among geometrically correlated neighbors. The core objective of our method is to discourage over-concentrated regions of the token space while preserving the model’s original scoring function. By redistributing probability mass away from highly crowded neighborhoods and toward less crowded alternatives, the proposed decoding strategy aims to mitigate geometric over-concentration during generation. Importantly, the method operates entirely at inference time and does not modify model parameters or require additional training.

% Motivated by the empirical association identified above, we hypothesize that excessive embedding-space crowding reflects a decoding regime in which probability mass is concentrated over geometrically correlated tokens, limiting effective exploration of alternative next-token choices.

% 受以上intuition的启发，我们进一步构建了

\subsection{Method Overview}

% 在每一个decoding步骤

In autoregressive generation, a language model predicts a next-token distribution over the vocabulary conditioned on the preceding context, and samples a token at each decoding step. % \method{} is a plug-in sampling-based method that reweights next-token distribution to alleviate embedding-space crowding, thereby promoting more robust and diverse reasoning trajectories. 
%\method{} reweights the next-token distribution to directly alleviate crowding. This plug-in sampling strategy promotes more robust and diverse reasoning trajectories.
Formally, at decoding step $t$, let $V$ denote the vocabulary and let $P(x_t \mid x_{1:t-1})$ denote the model's next-token distribution. Decoding under \method{} proceeds as follows:

\paragraph{Step 1: Selecting an effective correction set.} 
% We apply corrections only where embedding-space crowding is most likely.
To improve computational efficiency and to avoid unnecessary intervention, we apply corrections only where embedding-space crowding is most pronounced.
% To improve computational efficiency, we apply corrections only where embedding-space crowding is most pronounced.
Specifically, we restrict the correction to non-negligible tokens by defining
$S_t=\{\, i \mid p_{t,i} \geq \varepsilon \,\}$ at decoding step $t$,
where $\varepsilon$ is a fixed threshold.

% \noindent\textbf
\paragraph{Step 2: Computing token- and step-level crowding.}
% Following Definitions~3.1--3.2, we compute crowding within the effective correction set $S$. 
Following Definitions~3.1--3.2, we compute crowding within the effective correction set $S$, which gives us the specific crowding measures for the current tokens and the step.
For each $i \in S$ with probability $p_i$, token-level crowding is
\begin{equation}
\label{equ:token2}
\mathrm{Crowd}_t^{\text{token}}(i)
=
\sum_{j \in S_t \setminus \{i\}} p_{t,j} \, \bigl|\cos(e_i, e_j)\bigr| .
\end{equation}
We then compute $\mathrm{Crowd}^{\text{step}\dagger}(t)$, an adjusted (nonlinearly weighted) variant of step-level crowding, by aggregating token-level crowding at decoding step $t$:
\begin{equation}
\label{equ:adjustedstep}
\mathrm{Crowd}^{\text{step}\dagger}(t)
=
\sum_{i \in S_t}
p_{t,i} \, (e^{p_{t,i}} - 1)\, \mathrm{Crowd}_t^{\text{token}}(i).
\end{equation}

\paragraph{Step 3: Computing correction factors.}
%Combining the step-level adjusted crowding in Eq.~\ref{equ:adjustedstep} with the probability mass over $S$, we compute a step-level correction strength factor $\lambda_t$, where $\tau \in [0,1]$ controls the overall correction strength:
%The step-level correction strength factor $\lambda_t$, which controls the overall distribution adjustment, is computed by combining Eq.~\ref{equ:adjustedstep} with the probability mass over $S$. The hyperparameter $\tau \in [0, 1]$ modulates the global correction strength, where higher values of $\tau$ correspond to stronger correction:
We compute a step-level correction strength factor $\lambda_t$ to control the overall distribution adjustment. This factor combines the step-level adjusted crowding from Eq.~\ref{equ:adjustedstep} with the probability mass over $S$. The hyperparameter $\tau \in [0, 1]$ modulates the global correction strength, where higher values of $\tau$ correspond to stronger correction:
\begin{equation}
\lambda_t
=
\frac{\tau \sum_{i \in S_t} p_i}
{\mathrm{Crowd}^{\text{step}\dagger}(t)\,\bigl(1 - \tau \sum_{i \in S} p_i\bigr)} .
\end{equation}

%We then obtain the token-wise crowding-aware correction factor $\alpha_i$ for each $i \in S$:
We then combine this with Eq.~\ref{equ:token2} to obtain a token-wise crowding-aware correction factor $\alpha_i$ for each $i \in S$. This factor is used directly to adjust the next-token distribution, penalizing tokens with both higher probability and higher crowding:
\begin{equation}
\alpha_{t,i}
=
\frac{1}{1 + \lambda_t \, (e^{p_{t,i}} - 1)\, \mathrm{Crowd}_t^{\text{token}}(i)} .
\end{equation}

\paragraph{Step 4: Crowding-aware correction.}
For each $i \in S$, we reweight the token by applying $\alpha_i$, downweighting tokens that have both higher probability and a higher embedding-space crowding score:
\begin{equation}
\tilde{p}_{t,i} = \alpha_{t,i} \, p_{t,i} .
\end{equation}

\paragraph{Step 5: Renormalization and reinsertion.}
We rescale $\{\tilde{p}_{t,i}\}_{i \in S_t}$ to preserve the original mass $\sum_{i \in S_t} p_{t,i}$ and form the final distribution over $V$:
\begin{equation}
\label{equ:normalize}
p_{t,i}'
=
\begin{cases}
\tilde{p}_{t,i} \cdot \dfrac{\sum_{k \in S_t} p_{t,k}}{\sum_{k \in S_t} \tilde{p}_{t,k}}, & i \in S_t, \\[6pt]
p_{t,i}, & i \in V \setminus S_t .
\end{cases}
\end{equation}
All subsequent sampling is performed from $\{p_i'\}_{i \in V}$.

\subsection{Theoretical Intuition and Design Rationale}
% 动机
% embedding-space crowding 是一个在decoding阶段现象，reflects a decoding regime in which probability mass becomes concentrated over geometrically correlated tokens. 在section~\ref{sec:define_analysis}，我们基于定义并进行量化，在分析中，我们发现embedding-space crowding与decoding failures有着统计上的正相关。这种几何过度集中可能会使模型倾向于从标记空间的狭窄区域进行采样，从而限制推理路径的多样性并约束解码轨迹的演变。intuition是this structural bias contributes to brittle or incorrect generation outcomes, motivating the need for a crowd-aware correction mechanism during sampling.
% Building on Section~\ref{sec:define_analysis}, we find a statistically significant positive correlation between embedding-space crowding and decoding failures. 
Building on Section~\ref{sec:define_analysis}, we establish a statistically significant negative correlation between embedding-space crowding and reasoning success.
Embedding-space crowding is a decoding-time phenomenon where probability mass concentrates on geometrically correlated tokens. This over-concentration can bias sampling toward a narrow region of the token space. It can reduce trajectory diversity and restrict how decoding evolves, which may lead to brittle or incorrect generations. Motivated by this intuition, \method{} applies a crowd-aware correction mechanism during sampling to promote more robust and diverse reasoning trajectories.  
We now detail the design and rationale of its core components: (1) the effective correction set, (2) the token-level crowding-aware correction factor, (3) the step-level correction strength factor, and (4) renormalization.

\paragraph{Effective Correction Set Design.}
For efficiency and stability, we apply crowding-aware correction only to high-probability tokens in an effective correction set. 
% Crowding is most pronounced when probability mass concentrates on geometrically correlated tokens, while low-probability tail tokens have limited impact on decoding and incur unnecessary computation. 
Crowding is most pronounced on geometrically correlated, high-probability tokens. Correcting the low-probability tail is inefficient due to its limited impact on decoding, introducing unnecessary computation and intervention.
We therefore use $S=\{\, i \mid p_i \geq \varepsilon \,\}$ with a fixed threshold $\varepsilon$ (default $\varepsilon=10^{-2}$). This choice balances coverage of salient candidates and computational cost. We evaluate alternative thresholds in ablations.

\paragraph{Crowding-aware Correction Factor Design.}
% 校正因子$\alpha$ 依据每个 token 的概率大小及其在 embedding 空间中的拥挤程度进行调整，旨在在尊重模型原始预测分布的前提下，抑制空间中过度集中的区域。$\alpha$d的结构是一个乘性衰减函数。具体而言，对那些概率较高且在嵌入空间中重复性强的 token 会施加有力的下调；而对稀疏或低概率的 token 则几乎不做影响。指数项$(e^{p_i} - 1)$,引入了非线性权重，是一种连续平滑的调节机制。这种非线性设计强调在同一crowding下对高概率的token更强的惩罚，同时避免了对小$p_i$过度削弱。类似的结构也体现在 step-level adjusted crowding 值中\mathrm{Crowd}^{\text{step}\dagger}(t)。总体而言，这种设计使 \method{} 无需依赖额外的全局启发式或不确定性度量，即可有效抑制冗余但占主导地位的 token。
% The correction factor $\alpha$ adjusts each token based on its probability and its degree of crowding in the embedding space, aiming to suppress overly concentrated regions while respecting the model’s original predictive distribution. The structure of $\alpha$ is a multiplicative modulation function. Specifically, tokens with higher probabilities and stronger redundancy in embedding space are 更加 down-weighted, while sparser or lower-probability tokens are left 更 unaffected. The exponential term $(e^{p_i} - 1)$ introduces a smooth nonlinear weighting mechanism. This design emphasizes stronger penalization for high-probability tokens under similar crowding conditions, while avoiding excessive suppression of low-probability items. A similar structure 也在 the step-level adjusted crowding$\mathrm{Crowd}^{\text{step}\dagger}(t)$中. Overall, this design enables \method{} to effectively suppress dominant but redundant tokens without relying on additional global heuristics or uncertainty-based metrics.
The correction factor $\alpha_i$ modulates each token according to its probability and embedding-space crowding, suppressing overly concentrated regions while preserving the model’s distributional shape. It acts as a multiplicative down-weighting term. Specifically, tokens with higher probability and stronger embedding redundancy receive larger penalties, whereas lower-probability or less crowded tokens are minimally affected. The exponential term $(e^{p_i} - 1)$ introduces a smooth nonlinear weighting mechanism. This design increases penalization for high-probability tokens under comparable crowding, without overly suppressing lower-probability candidates. 
% We use the same nonlinear weighting in the step-level adjusted measure $\mathrm{Crowd}^{\text{step}\dagger}(t)$. 
Overall, this design enables \method{} to effectively suppress dominant but redundant tokens without relying on additional global heuristics or uncertainty-based metrics.

\paragraph{Correction Strength Factor Design.}
% 校正强度因子 $\lambda_t$ 是每一步解码中控制整体惩罚幅度的关键参数，决定了 \method{} 对当前分布的调整程度。其设计目标是，使得第三步修正的期望保持在S集之后*tau的上下周围，从而使得整体修正幅度保持可以保持相对稳定，从而适应不同分布及温度情况。其具体数学推导过程见appendix。λ 作为全局尺度因子，使crowding修正过程具备可控性和稳定性。tau是可以调控的参数，调整修正的幅度，更大的tau幅度越大反之越小，我们的ablation study讨论了不同tau下的而结果。
% 校正强度因子 $\lambda_t$ 是每一步解码中控制整体惩罚幅度的关键参数，决定了 \method{} 对当前分布的调整程度。其设计目标是，使得第三步中修正的the expected total reduction在原始高概率候选集 $S$ 的总质量 $\sum_{i \in S} p_i$ 附近轻微上下浮动，确保整体校正幅度的可控性与稳定性，并适应不同的分布形态与温度设定。$\tau \in [0,1]$ 是一个可调超参数，用于控制整体修正幅度。较大的 $\tau$ 值会产生更强的修正，而较小的值则能更紧密地保留原始分布。关于 $\lambda_t$ 的数学构造细节与推导过程可参见附录。我们在消融实验中进一步讨论了不同 $\tau$ 设置下的性能表现。
% 校正强度因子 $\lambda_t$ 是每一步解码中控制整体惩罚幅度的关键参数，决定了 \method{} 对当前分布的调整程度。其设计目标是，使得修正后分布 ${\tilde{p}i}{i \in S}$ 与原始分布 ${p_i}{i \in S}$ 的总质量差约为 $\tau \sum{i \in S} p_i$，即整体削减幅度占原始候选集的 $\tau$ 倍。从而确保整体校正幅度的可控性与稳定性，并适应不同的分布形态与温度设定。$\tau \in [0,1]$ 是一个可调超参数，用于控制整体修正幅度。较大的 $\tau$ 值会产生更强的修正，而较小的值则能更紧密地保留原始分布。关于 $\lambda_t$ 的数学构造细节与推导过程可参见附录。我们在消融实验中进一步讨论了不同 $\tau$ 设置下的性能表现
At decoding step $t$, the step-level strength factor $\lambda_t$ controls the overall correction magnitude applied by \method{} to the current distribution. We choose $\lambda_t$ so that the total probability mass removed from the candidate set $S$ is approximately $\tau \sum_{i \in S} p_i$, which reduces the original mass over $S$ by a fraction $\tau$. This ensures stable and interpretable control and adapts across different distribution shapes and temperature settings. The hyperparameter $\tau \in [0,1]$ tunes the overall strength. Larger $\tau$ applies a stronger correction, while smaller $\tau$ better preserves the original distribution. We provide the derivation of $\lambda_t$ in the detailed Appendix~\ref{appendix:derivation}.

%and study the effect of $\tau$ in ablations.

\paragraph{Renormalization.}
We rescale the penalized probabilities $\{\tilde{p}_{t,i}\}_{i \in S_t}$ to preserve the original total mass of the correction set, $\sum_{i \in S_t} p_{t,i}$. These rescaled values are then merged back into the full vocabulary distribution.
%This design ensures that tokens with higher initial probability and higher crowding receive stronger penalties, causing their probabilities to decrease after renormalization. 
% This design ensures that tokens with higher initial probability and crowding, having received stronger penalties during the correction phase, end up with lower probabilities after renormalization. 
% Tokens with higher initial probability and crowding receive stronger penalties during the correction phase. This design ensures that their probabilities are lower after the subsequent renormalization.
Tokens with higher initial probability and crowding are penalized more heavily during correction, resulting in lower probabilities after renormalization. 
Conversely, tokens less affected by crowding are relatively boosted. Thus, probability mass is systematically shifted away from high-crowding, high-probability regions.
Furthermore, by restricting modifications to the effective candidate set $S_t$, we avoid unnecessary perturbations to the long-tail tokens, which minimizes the introduction of additional noise and randomness.

% ADVANTAGES OF \method{} FOR SAMPLING-BASED DECODING
%Unlike temperature-based or entropy-aware sampling, which reshape the distribution using uncertainty or global heuristics, our approach exploits local embedding-space geometry during decoding. Temperature uniformly flattens or sharpens the distribution, and entropy-aware methods adapt based on global dispersion. In contrast, \method{} selectively down-weights tokens that are both high-probability and geometrically redundant. This yields targeted, structure-aware corrections without additional model passes or external signals.
\subsection{Advantages of \method{} for Sampling}
\method{} introduces a crowding-aware reweighting mechanism for sampling-based decoding with several advantages. % First, \method{} leverages geometric structure among token embeddings to reflect the internal organization of the next-token distribution. 
%First, \method{} leverages the geometric structure of token embeddings to capture the next-token distribution's internal organization.
%\method{} introduces a crowding-aware reweighting mechanism for sampling-based decoding. It exploits the geometric proximity structure in the token embedding space to estimate local crowding around candidate tokens and reweight the next-token distribution accordingly.
% In contrast, existing truncation- and reweighting-based methods mostly adjust global cutoff or temperature using fixed scalars, uncertainty measures, or other distributional statistics, typically ignore token--token geometric relations. 
%In contrast, truncation- and temperature-based methods%~\cite{holtzman2019curious,ackley1985learning} 
%adjust global cutoffs or temperature, applying coarse-grained adjustments to the overall distribution.
First, \method{} leverages the geometric structure of token embeddings to capture the internal organization of the next-token distribution. In contrast, truncation- and temperature-based methods operate solely on probability values, applying global, coarse-grained adjustments to the overall distribution.
% Second, \method{} performs targeted, localized reweighting rather than global reshaping. It suppresses embedding-redundant candidates that are highly similar in embedding space, which more directly improves reasoning robustness and diversity. 
% Second, \method{} performs targeted, localized reweighting instead of global reshaping, focusing on high-probability and embedding-redundant candidates.
% By comparison, temperature-based schemes promote exploration by globally flattening the distribution, often at the cost of increased randomness. 
Second, \method{} performs localized, geometry-aware reweighting of the next-token distribution, focusing correction on high-probability tokens in crowded embedding regions. By comparison, temperature-based methods%~\cite{zhu2024hot}
promote exploration by uniformly flattening the distribution, often at the cost of increased randomness.
Third, \method{} operates on the single-step distribution and requires no extra forward passes, rollouts, or external signals. Finally, as a plug-in probability modifier, \method{} can integrate with standard samplers like temperature and top-$p$, enabling easy deployment.

\begin{table*}[htbp]
\caption{The results of our main experiments under the two configurations. \method{} enhances both robustness and diversity in reasoning. The metric Dist-n and SemDiv correspond to Distinct-n and Semantic Diversity. Across datasets and both decoding configurations, \method{} generally improves reasoning performance and diversity relative to standard sampling.}
    \centering
    \small
    \setlength{\tabcolsep}{0.8mm}
    \begin{tabular}{lcccccccccccc}
\toprule 
\multicolumn{1}{c}{} & \multicolumn{4}{c}{\textbf{AIME24}} & \multicolumn{4}{c}{\textbf{AIME25}} & \multicolumn{4}{c}{\textbf{HMMT25}} \\
\cmidrule(lr){2-5} \cmidrule(lr){6-9} \cmidrule(lr){10-13}
 & \textbf{Avg@32}  & \textbf{Pass@8}  & \textbf{Dist-N} & \textbf{SemDiv} & \textbf{Avg@32}  & \textbf{Pass@8}  & \textbf{Dist-N} & \textbf{SemDiv} & \textbf{Avg@32}  & \textbf{Pass@8}  & \textbf{Dist-N} & \textbf{SemDiv} \\

\midrule
    \rowcolor{gray!20}
    \multicolumn{13}{c}{\emph{\textbf{Temp = 1.0, Top-p = 1.0}}} \\

\noalign{\vskip 3pt}
Standard Sampling  &  33.75 & 55.07 & 51.00  & 15.17 &  35.94 & 57.33 & 50.36  & 16.61 & 13.85 & 27.74 & 51.01  & 14.78  \\
%\cdashline{1-15} 
%\noalign{\vskip 3pt}
% \quad + \method{} (Ours)  &  34.06 & 56.59 & 51.91 & 15.25 & 36.46 & 58.91 & 52.03 & 17.70 & 14.58 & 30.58 & 51.93 & 15.46 \\
\quad + \method{} (Ours)  &  \textbf{34.06} & \textbf{56.59} & \textbf{51.91} & \textbf{15.25} & \textbf{36.46} & \textbf{58.91} & \textbf{52.03} & \textbf{17.70} & \textbf{14.58} & \textbf{30.58} & \textbf{51.93} & \textbf{15.46} \\
\midrule
    \rowcolor{gray!20}
    \multicolumn{13}{c}{\emph{\textbf{Temp = 0.7, Top-p = 0.95}}} \\
\noalign{\vskip 3pt}
Standard Sampling  & 34.17	& \textbf{57.53}  &  44.70 & 12.43  
& 35.73 & 58.69 &  43.18 & \textbf{14.71} & 11.88 & 21.44 & 41.53 & 12.02 \\
%\cdashline{1-15} 
%\noalign{\vskip 3pt}
\quad + \method{} (Ours)  &  \textbf{34.58}	& 56.16
  & \textbf{44.76} & \textbf{12.52} & \textbf{37.29} & \textbf{60.37} & \textbf{44.46} & 14.50 & \textbf{12.60}
 & \textbf{24.43} & \textbf{42.28} & \textbf{12.10} \\

\bottomrule
\end{tabular}
    \label{tab:main}
\end{table*}

\subsection{Implementation Details}

% IMPLEMENTATION DETAILS
% Integration into Decoding Pipelines.
% token embedding获得（白盒模型），token embedding的计算（矩阵）
% Parameter Selection Guidelines. 1.Choosing the tau epison; 2.Combining with Other Techniques
% 效率：真实时间比例。crowding计算的复杂度，理论上。

\paragraph{Integration into Decoding Pipelines.}
%Our method is implemented as a post-softmax reweighting module and can be seamlessly integrated into existing sampling-based decoding pipelines. It requires access to the model's token embedding matrix, which is available in all standard causal language models (e.g., via the final embedding layer tied to the output projection).
Designed for easy adoption, \method{} acts as a lightweight, post-softmax reweighting step. It can be integrated into standard decoding pipelines without any fine-tuning or architectural changes. As it relies solely on the model's existing token embeddings and the next-token logits, it is compatible with any open-source LLM. Crowd-aware sampling is also compatible with common sampling methods such as top-$k$, top-$p$, and temperature scaling. In practice, \method{} can be inserted between the temperature scaling process and any subsequent filtering step.

\paragraph{Token Embedding Access and Computation.}
%At each decoding step, we retrieve the embedding vectors for all candidate tokens in the correction set $S$. We compute token-level crowding via cosine similarity. This is implemented efficiently as a matrix multiplication followed by row-wise operations, taking advantage of GPU parallelism.
At each decoding step, we use the static embedding vectors for all candidate tokens in the correction set $S_t$. The token-level crowding score is computed based on pairwise cosine similarities. The scores are computed in two vectorized steps. A pairwise similarity matrix is derived from the candidate embeddings, followed by a row-wise aggregation to produce the final token-level crowding measure. Since the effective correction set $S_t$ is limited in size (typically $|S_t| \leq 100$ for $\epsilon=10^{-2}$).%, the required overhead remains small. 
Its computation, along with the subsequent aggregation, amounts to negligible overhead and is implemented with standard, vectorized tensor operations.

%\paragraph{Parameter Selection Guidelines.}
%We set the minimum probability threshold $\varepsilon = 0.01$ to filter out low-probability tokens. The correction strength parameter $\tau$ 一般在$[0.2,0.5]$ 之间，来controls how aggressively the distribution is reshaped. A detailed 内容 is provided in Section~\ref{sec:exp} and Appendix~\ref{}.
%\paragraph{Parameter Selection Guidelines.}
%We set the minimum probability threshold $\varepsilon = 0.01$ to filter out low-probability tokens. The correction strength parameter $\tau$, which controls how aggressively the distribution is reshaped, is typically set within $[0.2, 0.5]$. More details are provided in Section~\ref{sec:exp} and Appendix~\ref{}.

%\paragraph{Compatibility with Other Decoding Techniques.}
%Crowd-aware sampling is orthogonal to common sampling-based method such as top-$k$, top-$p$, and temperature scaling. In practice, it can be applied after those filters and before sampling, as a final geometry-aware reweighting step.
%In practice, \method{} can be inserted between standard temperature scaling and any subsequent filtering step.

%\paragraph{Efficiency and Overhead.}
%Our implementation introduces minimal computational overhead. On a 1.7B-parameter model (Qwen-1.7B), the average per-token decoding latency increases by approximately 10--15\%. Theoretical complexity scales as $O(k^2)$ with the number of retained tokens $k = |S|$, which is typically $\leq 100$ after filtering.

\section{Experiments}
\label{sec:exp}
\subsection{Experiments Setup}
% Our experiments were performed using 在推理任务中代表性的模型Qwen3-1.7B, Qwen3-4B~\cite{}, and Hunyuan-1.8B-Instruct~\cite{}。数学推理是最具代表性的推理任务，困难且可验证。我们在三个数学推理数据集上进行实验，AIME24t~\cite{}、AIME25t~\cite{}、HMMT25t~\cite{}。我们分别在不同temperture和top-p情景下做实验，包括temperture=1.0、top-p=1.0，temperture=0.7、top-p=0.95，两个常见设置。For evaluation，we repeat the evaluation set for 32 times and report avg@32 for results stability and pass@8 for……。avg@32是mean，pass@8我们跟随……。
\noindent\textbf{Models.} 
 % 我们的主实验在Qwen3-1.7B模型上进行。Qwen3-1.7B是一个有代表性的
% Qwen3 Technical Report
We adopt Qwen3-1.7B~\cite{yang2025qwen3} as our backbone, as it offers a favorable tradeoff between capability and computational cost and has been used in several recent studies~\cite{ulli2025medqwen,li2025repo}.
We further evaluate our method on Qwen3-4B~\cite{yang2025qwen3} and Hunyuan-1.8B-Instruct~\cite{tencent_hunyuan_18b_instruct} to assess its robustness and generalizability across different backbone models.

% \noindent\textbf{Benchmarks.} Mathematical reasoning datasets, with well-defined problems and verifiable answers, provide an ideal benchmark for evaluating reasoning capability. We evaluate \method{} sampling on three typical math reasoning benchmarks: AIME24~\cite{}, AIME25~\cite{}, and HMMT25~\cite{}.

\noindent\textbf{Benchmarks.} Mathematical reasoning datasets feature well-defined problems with verifiable answers, making them an ideal testbed for assessing complex reasoning capability. We evaluate \method{} on three representative benchmarks: AIME24~\cite{aime2024}, AIME25~\cite{aime2025}, and HMMT25~\cite{hmmt2025}.

\noindent\textbf{Baselines and Hyperparameters.} 
% 我们使用主流的decoding场景作为baseline，包括top-p、temperature的场景，将\method{}和未使用\mehtod{}做对比。
%We adopt standard decoding settings as our baseline (i.e., top-p nucleus sampling with temperature) and compare \method{} against the same setup without \method{}.
We adopt standard decoding settings as our baseline (top-$p$ nucleus sampling with temperature) and compare \method{} against the same setup without \method{} to control for decoding settings.
Experiments were conducted under two common sampling configurations in our main experiments: (1) $\text{temperature}=1.0, \text{top-}p=1.0$ and (2) $\text{temperature}=0.7, \text{top-}p=0.95$. 
%Other experiments set in configuration (1).
All other experiments on different models use the same setup as configuration (1).
In the \method{} hyperparameters, setting $\varepsilon$ to 0.01 and %$\tau$ between 0.2 and 0.5 
$\tau \in \{0.2, 0.3, 0.4, 0.5\}$ leads to good performance. For more details, please refer to Appendix~\ref{app:par}.

\noindent\textbf{Evaluation Metrics.}
For each benchmark, we replicate the evaluation set 32 times and report avg@32, pass@8, Distinct-$n$, and semantic diversity. Avg@32 is the mean score over 32 repeated runs, indicating result stability. Pass@8 is the fraction of problems solved in at least one of 8 independent trials, following~\cite{chen2021evaluating}. Distinct-n~\cite{jiweidiversitypromoting} (with $n=4$) is the ratio of unique n-grams to total n-grams in the generated set, measuring diversity in the form of language. 
Semantic diversity is computed from embeddings using all-MiniLM-L12-v2 model~\cite{st_allminilm_l12_v2}. We define $m_{\text{div}} = 1 - \frac{1}{|\mathcal{P}|}\sum_{(i,j)\in \mathcal{P}} \mathrm{cos}(z_i, z_j)$, where $\mathcal{P}$ is the set of all unordered pairs and $z_i$ denotes the embedding of sample $i$~\cite{TevetEvaluating}. For efficiency and context-length constraints, semantic diversity uses only the first 512 tokens of each sample's final output. For presentation in tables, this score is scaled by 100.

\begin{table}[htbp]
\caption{The results of experiments on Qwen3-4B model under the configuration (1). \method{} delivers consistent or improved on avg@32 and diversity metrics across benchmarks.}
    \centering
    \small
    \setlength{\tabcolsep}{1.2mm}
    \begin{tabular}{lcccc}
\toprule 
& \textbf{Avg@32}  & \textbf{Pass@8}  & \textbf{Dist-N} & \textbf{SemDiv} \\

\midrule
\rowcolor{gray!20}
    \multicolumn{5}{c}{\textbf{AIME24}} \\
\noalign{\vskip 3pt}

Standard Sampling    &  50.42 & \textbf{60.96}  &  51.14 &14.99      \\
%\cdashline{1-15} 
%\noalign{\vskip 3pt}
\quad + \method{} (Ours)  &  \textbf{50.63}	& 60.68  &  \textbf{54.60} & \textbf{16.30}   \\
\midrule
\rowcolor{gray!20}
    \multicolumn{5}{c}{\textbf{AIME25}} \\
\noalign{\vskip 3pt}
Standard Sampling    &  65.00 & 81.79  &  52.37 & 16.30    \\
%\cdashline{1-15} 
%\noalign{\vskip 3pt}

\quad + \method{} (Ours)  &  \textbf{65.21} & \textbf{82.47} &  \textbf{55.34} & \textbf{17.36}  \\
\midrule
\rowcolor{gray!20}
    \multicolumn{5}{c}{\textbf{HMMT25}} \\
\noalign{\vskip 3pt}
Standard Sampling    &  23.75 &	31.82  & 51.56 &  14.78  \\
%\cdashline{1-15} 
%\noalign{\vskip 3pt}
\quad + \method{} (Ours)  &  \textbf{25.31} &	\textbf{33.75}  & \textbf{54.17} & \textbf{15.10} \\
%\midrule

%(25.31-23.75)
%(33.75-31.82)
\bottomrule
\end{tabular}
    \label{tab:other}
\end{table}

\subsection{Results}
\paragraph{Main Results.} 
% 我们在主实验中使用qwen3-1.7B模型在两个configurations下进行实验，结果如表~\label{tab:main}.在两个configurations下，\method{}对于准确率和多样性均有稳定的提升。其中，在temperature=1.0, top-p=1.0，在三个数据集中，avg@32平均提升了0.52个百分点, Pass@8平均提升了1.98个百分点,distinct-n平均提升了1.17个百分点,semantic diversity平均提升了0.62个百分点. 说明通过\method{}，通过修正embedding-space crowding后，模型推理的robustness和diversity均有提升。
% For our main experiments, we evaluate the Qwen3-1.7B model under two configurations, with results presented in Table~\ref{tab:main}. Under both settings, \method{} consistently improves both accuracy and diversity. Specifically, with $\text{temperature}=1.0, \text{top-}p=1.0$, our method achieves an average gain of 0.52 percentage points in avg@32, 1.98 points in pass@8, 1.17 points in distinct‑n, and 0.62 points in semantic diversity across the three datasets. With $\text{temperature}=0.7, \text{top-}p=0.9$, our method achieves an average gain of 0.90 percentage points in avg@32, 1.10 points in pass@8, 0.70 points in distinct‑n, while average semantic diversity remains nearly unchanged across the three datasets. These results demonstrate that \method{} enhances both the robustness and diversity of model inference by mitigating embedding‑space crowding. 
For our main experiments, we evaluate Qwen3-1.7B under two decoding configurations, as shown Table~\ref{tab:main}. Across datasets and settings, \method{} yields consistent gains in Avg@32 and improves overall pass rate and diversity metrics on average. With $\text{temperature}=1.0$ and $\text{top-}p=1.0$, \method{} improves Avg@32 by +0.52 points and Pass@8 by +1.98 points on average, while also increasing Dist-N (+1.17) and semantic diversity (+0.62). Under a more conservative setting with $\text{temperature}=0.7$, $\text{top-}p=0.95$, \method{} continues to improve Avg@32 (+0.90) and Pass@8 (+1.10) on average and modestly increases Dist-N (+0.70), while semantic diversity remains essentially unchanged (-0.01 on average). 
These results show that \method{} generally improves reasoning performance and diversity relative to standard sampling across datasets and both decoding configurations.
% Notably, the largest Pass@8 gains are observed on the more challenging HMMT25 benchmark (up to +2.99), suggesting that mitigating embedding-space crowding is particularly beneficial when the next-token distribution is more geometrically concentrated.
% Notably, the largest Pass@8 gains are observed on HMMT25 (up to +2.99), indicating that \method{} has the potential to be especially helpful on harder benchmarks.

Beyond the averaged gains, we observe that the benefits of \method{} depend on the decoding configurations. Diversity improvements are more pronounced at $T{=}1.0$, $top-p{=}1.0$, while remaining modest under $T{=}0.7$, $top-p{=}0.95$ with semantic diversity largely unchanged. This indicates that the diversity gains from \method{} are consistent across decoding configurations. They are larger when sampling permits broader exploration, and remain measurable under tighter constraints. Notably, Avg@32 improves across all datasets in both configurations, suggesting a more reliable overall generation quality even when individual metrics may occasionally fluctuate. 
Furtermore, the largest Pass@8 gains are observed on the more challenging HMMT25 benchmark (up to +2.99), indicating that \method{} has the potential to be especially helpful on harder benchmarks.
%We also see the largest Pass@8 gains on HMMT25, suggesting that \method{} may be especially helpful on harder benchmarks where standard sampling performs less reliably.
% We also find that the largest Pass@8 improvements occur on HMMT25, indicating that \method{} may be particularly beneficial in harder regimes where standard sampling is more failure-prone.

% \paragraph{Experiments on Other Model.} 
% 0.253125	0.337454123	0.2375	0.318214014
% 我们进一步验证我们方法在其他模型上的性能，包括Qwen3-4B模型和hunyuan-1.8B模型，结果如表~\label{tab:other_models}. 在两个模型上，\method{}获得到了更优或者近似的avg32和pass@8，以及持续更优的distinct‑n和semantic diversity数值。These results demonstrate that \method{} 在不同模型上均可以获得更优或稳定的robustness且更优的diversity。
% We further evaluate our method on other models, including Qwen3-4B and Hunyuan-1.8B. The results are shown in Table~\ref{tab:other_models}. Across both models, \method{} achieves superior or comparable scores in avg@32 and Pass@8, while consistently yielding higher distinct‑n and semantic diversity. These results demonstrate that \method{} can attain improved or stable robustness along with enhanced diversity across different models.

\begin{table}[htbp]
\caption{The results of experiments on Hunyuan-1.8B-Instruct model under the configuration (1) on AIME24. 
%\method{} delivers a substantial accuracy gain, with a slightly decrease in diversity metric.
%On Hunyuan-1.8B-Instruct (AIME24), 
\method{} delivers clear accuracy improvements, with small declines in diversity metrics
}
    \centering
    \small
    \setlength{\tabcolsep}{1.4mm}
    \begin{tabular}{lcccc}
\toprule 
& \textbf{Avg@32}  & \textbf{Pass@8}  & \textbf{Dist-N} & \textbf{SemDiv} \\

\midrule
%\rowcolor{gray!20}
%    \multicolumn{5}{c}{\textbf{AIME24}} \\
%\noalign{\vskip 3pt}

% 23.54 & 39.94	22.18 & 35.01

Standard Sampling    &  22.18 & 35.01  &  \textbf{81.54} & \textbf{19.27}      \\
%\cdashline{1-15} 
%\noalign{\vskip 3pt}
\quad + \method{} (Ours)  &  \textbf{23.54} & \textbf{39.94}  & 79.41  & 18.81   \\

%(25.31-23.75)
%(33.75-31.82)
\bottomrule
\end{tabular}
    \label{tab:other2}
\end{table}

\paragraph{Experiments on Other Models.}
To examine transferability, we additionally evaluate \method{} on a larger Qwen3-4B model and Hunyuan-1.8B-Instruct as a non-Qwen model.  As shown Table~\ref{tab:other}, on Qwen3-4B, \method{} improves Avg@32 across all three benchmarks and increases diversity metrics, both Dist-N and SemDiv, consistently, while Pass@8 improves on AIME25 and HMMT25 with a minor decrease on AIME24 (-0.28). As shown Table~\ref{tab:other2},
% \method{} yields a substantial accuracy gain, both Avg@32 (+1.36) and Pass@8 (+4.93) in AIME24, though Dist-N and SemDiv slightly decrease. 
on Hunyuan-1.8B-Instruct with AIME24 benchmark, \method{} improves accuracy, increasing Avg@32 (+1.36) and Pass@8 (+4.93), though Dist-N and SemDiv slightly decrease. 
Importantly, the fraction of near-exact semantic matches used in SemDiv computation (similarity $>0.999$) drops from 1.04\% to 0.39\%, indicating reduced near-duplicate outputs despite the small change in aggregate diversity scores. Overall, these results suggest that \method{} generalizes beyond the base model, while the diversity-accuracy tradeoff can be model-dependent.

\subsection{Ablation Study}
% 我们进一步分析\method{}中component的合理性，我们以主实验模型qwen3-1.7B和AIME25数据集为代表进行……的分析。
To further validate the design of \method{}, we conduct an ablation study using the Qwen3-1.7B model and the AIME25 dataset as representative cases.
% 其中，主要包括nonlinear weighting mechanism和step-level correction strength factor两个部分。
Our analysis focuses on two key components: the nonlinear weighting mechanism and the step-level correction strength factor.

\begin{table}[htbp]
\caption{Ablation on nonlinear and linear weighting in \method{}. Both improve over standard sampling, supporting the effectiveness of crowding-aware correction, with complementary accuracy-diversity profiles.
}
    \centering
    \small
    \setlength{\tabcolsep}{0.8mm}
    \begin{tabular}{lcccc}
\toprule 
& \textbf{Avg@32}  & \textbf{Pass@8}  & \textbf{Dist-N} & \textbf{SemDiv} \\

\midrule

% 23.54 & 39.94	22.18 & 35.01

Standard Sampling    &  35.94 & 57.33 & 50.36  & 16.61      \\
%\cdashline{1-15} 
%\noalign{\vskip 3pt}
\quad + \method{} (Nonlinear)  &  36.46 & 58.91 & \textbf{52.03} & \textbf{17.70}   \\
%  35.94 & 57.33 & 50.36  & 16.61 
% \textbf{36.46} & \textbf{58.91} & \textbf{52.03} & \textbf{17.70}
%\cdashline{1-15} 
%\noalign{\vskip 3pt}
\quad + \method{} (Linear)  &  \textbf{39.37} & \textbf{63.47}  & 50.97 & 17.31   \\
%(25.31-23.75)
%(33.75-31.82)

\bottomrule
\end{tabular}
    \label{tab:ab1}
\end{table}

\paragraph{Nonlinear Weighting Mechanism.} 
% 它的构建是使用一个exponential term $(e^{p_i} - 1)$，来引入一个平稳函数，稍微放大对于高概率的惩罚。为了验证其有效性，我们将$(e^{p_i} - 1)$替换成线性$p_i$，实验结果如表~\ref{}所示。实验发现，替换后其正确性和多样性均有下降，说明该项的有效性。
An exponential term, $(e^{p_{t,i}} - 1)$, is used to construct a smooth penalty function that slightly amplifies the correction for high-probability tokens. As shown in Table~\ref{tab:ab1}, replacing $(e^{p_{t,i}}-1)$ with a linear term $p_i$ still improves over standard sampling. This indicates that the main benefit comes from the crowding-aware correction itself, rather than a specific weighting choice. The two weightings yield different accuracy–diversity profiles. The linear form achieves higher Avg@32 (39.3) and Pass@8 (63.47). The exponential form yields higher diversity, both Dist-N (52.03) and SemDiv (17.70). These results suggest that the nonlinear and linear weighting modulates how strongly probability mass is shifted away from high-probability tokens. Nonlinear reweighting promotes exploration and diversity, but may increase decoding stochasticity. We leave a systematic characterization of this tradeoff, and adaptive weighting schemes, to future work.

%As shown in Table ~\ref{tab:ab1}, the model with the linear term shows a decline in both accuracy and diversity, confirming the importance of the exponential formulation.

\paragraph{Correction Strength Factor.} 
% the step-level strength factor $\lambda_t$ 依据每个decoding step时的整体概率分布 the overall correction magnitude，保证修正幅度的稳定。如表~\ref{}所示，当我们使用固定的strength factor时，具体而言固定为10，整体结果其正确性和多样性均有下降。此外，$\lambda_t$可以动态调整稳定性的derivation在appendix给出。
The step-level strength factor $\lambda_t$ adapts the correction magnitude to each step’s probability distribution for stability. Fixing $\lambda_t=10$ degrades accuracy.
% The step-level strength factor $\lambda_t$ adapts the overall correction magnitude based on the probability distribution at each decoding step, ensuring stable adjustment. 
%Replacing $\lambda_t$ with a fixed value of 10 results in degraded accuracy and diversity Replacing $\lambda_t$ with a fixed value of 10 results in degraded accuracy and diversity. %(Appendix~\ref{}).
%, as shown in Table ~\ref{}. 
The detailed results and derivation of $\lambda_t$ 
%for dynamical stability adjustment 
are provided in the Appendix~\ref{app:str} and ~\ref{appendix:derivation}.

\begin{figure}[ht]
  \begin{center}
    \centerline{\includegraphics[width=\columnwidth]{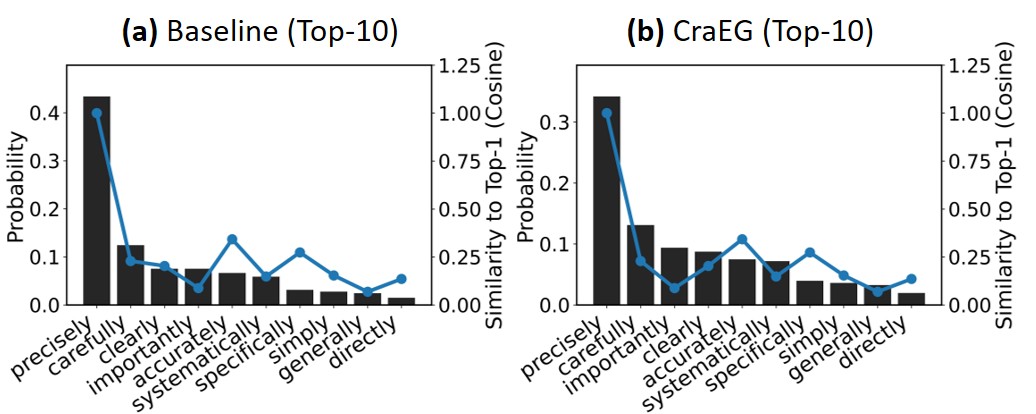}}
    \caption{
Step-level example of probability reallocation. Top-10 next-token probabilities and cosine similarity to the top-1 candidate (baseline vs. \method{}).}
    \label{fig:ana1}
  \end{center}
\end{figure}
\begin{figure}[ht]
  \begin{center}
    \centerline{\includegraphics[width=\columnwidth]{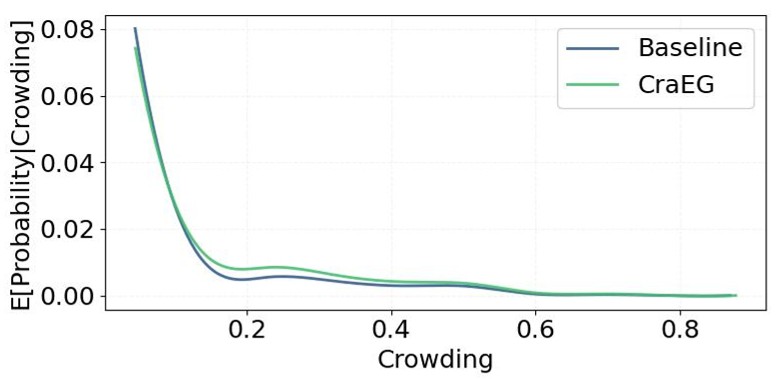}}
    \caption{
Trajectory-level token crowding analysis (top-30 per step). We report $\mathbb{E}[p \mid \text{crowding}]$ and the mean crowding aggregated over decoding steps. \method{} reduces the mean from 0.1934 to 0.1864.}
    \label{fig:ana2}
  \end{center}
\end{figure}

\subsection{Case Analysis}

%\paragraph{Step-level redistribution.}
%To provide intuitive insights into how \method{} reshapes decoding locally, we visualize the top-$K$ next-token distribution at representative steps (Figure~\ref{fig:case_step}). Compared to standard sampling, \method{} reduces top-1 dominance and redistributes probability mass across multiple plausible alternatives within the local candidate set. \textbf{Insight:} \method{} acts as a mild debiasing of overly peaked steps, yielding a flatter high-probability region without requiring a hard constraint on the candidate set.

\paragraph{Step-level redistribution.}
To provide intuitive insights into how \method{} reshapes decoding locally, we visualize the top-10 next-token distribution at a representative decoding step (Figure~\ref{fig:ana1}). Compared to standard sampling, \method{} assigns slightly different probabilities across the same top-10 candidates, shifting mass from the most dominant option to several close alternatives. \textbf{Insight:} the correction refines the model's preference among the same top-10 candidates, giving more room to plausible alternatives.

\paragraph{Trajectory-level crowding statistics.}
To provide complementary insights at the sequence level, Figure~\ref{fig:ana2} aggregates token-level crowding over the top-30 candidates along the full decoding trajectory and reports $\mathbb{E}[p \mid \text{crowding}]$. Compared to the baseline, \method{} assigns slightly higher expected probability in the low-to-moderate crowding regime, while the extreme regions remain similar. Over the entire trajectory, the mean top-30 token-level crowding decreases from 0.1934 (baseline) to 0.1864 (\method{}). \textbf{Insight:} \method{} shifts probability mass toward candidates with lower crowding in a gradual, trajectory-wide manner.
%\paragraph{Trajectory-level crowding statistics.}
%To provide complementary insights at the sequence level, Figure~\ref{fig:case_traj} aggregates token-level crowding over the top-30 candidates across all decoding steps and reports $\mathbb{E}[p\mid\text{crowding}]$. CraEG slightly increases expected probability in the low-to-moderate crowding range while leaving extreme regions largely unchanged, indicating a systematic but smooth shift in where probability mass is placed. Consistently, the mean top-30 token-level crowding decreases from 0.1934 (baseline) to 0.1864 (\method{}).

\section{Conclusion}
We take a geometry-based view of LLM decoding and show that embedding-space crowding influences how probability mass concentrates among localized region. Leveraging this perspective, we quantify crowding as a diagnostic signal to analyze how decoding-time crowding is associated with success on complex reasoning. We then propose \method{}, a lightweight crowding-aware sampling method that smoothly reweights next-token probabilities to improves the quality-diversity tradeoff. \method{} improves accuracy and generally strengthens diversity, with case studies revealing consistent trajectory-level shifts toward lower-crowding candidates. These results suggest that incorporating representation geometry into sampling can improve inference robustness and diversity, and motivate future geometry-aware decoding methods.
%


\section*{Impact Statement}

This paper presents work whose goal is to advance the field of machine learning by improving decoding for language models. The societal implications are similar to those of other methods that increase the quality and diversity of generated text: they can benefit downstream applications but may also amplify misuse if deployed without appropriate safeguards. We do not foresee impacts beyond these well-established considerations.

\nocite{langley00}

\bibliography{example_paper}
\bibliographystyle{icml2026}

%%%%%%%%%%%%%%%%%%%%%%%%%%%%%%%%%%%%%%%%%%%%%%%%%%%%%%%%%%%%%%%%%%%%%%%%%%%%%%%
%%%%%%%%%%%%%%%%%%%%%%%%%%%%%%%%%%%%%%%%%%%%%%%%%%%%%%%%%%%%%%%%%%%%%%%%%%%%%%%
% APPENDIX
%%%%%%%%%%%%%%%%%%%%%%%%%%%%%%%%%%%%%%%%%%%%%%%%%%%%%%%%%%%%%%%%%%%%%%%%%%%%%%%
%%%%%%%%%%%%%%%%%%%%%%%%%%%%%%%%%%%%%%%%%%%%%%%%%%%%%%%%%%%%%%%%%%%%%%%%%%%%%%%
\newpage
\appendix
\onecolumn
%\section{Other dataset}
%\label{app:other_d}

%You can have as much text here as you want. The main body must be at most $8$
%pages long. For the final version, one more page can be added. If you want, you
%can use an appendix like this one.

%The $\mathtt{\backslash onecolumn}$ command above can be kept in place if you
%prefer a one-column appendix, or can be removed if you prefer a two-column
%appendix.  Apart from this possible change, the style (font size, spacing,
%margins, page numbering, etc.) should be kept the same as the main body.

\section{Step-level ECDF analysis}
\begin{figure}[ht]
  \begin{center}
    \centerline{\includegraphics[width=0.5\columnwidth]{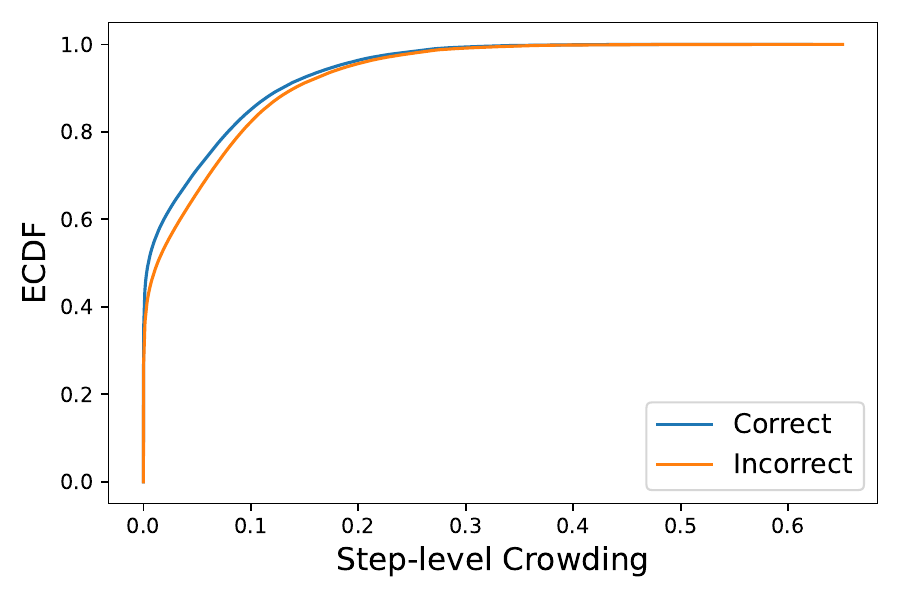}}
    \caption{ECDFs of step-level embedding-space crowding.
Empirical cumulative distribution functions of step-level crowding for decoding steps from correct and incorrect samples. Steps from incorrect samples are generally right-shifted toward higher crowding, consistent with elevated crowding being broadly present across steps rather than confined to a few extremes.}
    \label{fig:ecdf}
  \end{center}
\end{figure}

We provide additional step-level evidence complementing the sequence-level results in Section~\ref{sec:ea}. We pool all decoding steps from the generated samples and group them by the final correctness of the parent sample. Figure~\ref{fig:ecdf} plots the empirical cumulative distribution functions (ECDFs) of step-level crowding for steps from correct and incorrect samples. The incorrect ECDF is generally right-shifted toward higher crowding across most quantiles. Since steps within a sample are not independent, we report this as a descriptive comparison. The observed shift is consistent with elevated crowding being broadly present across steps rather than confined to a small number of extreme steps.

\section{Additional Regression Analysis}
\label{app:regression}
We report additional logistic regression results analyzing the relationship between embedding-space crowding, entropy, and reasoning correctness. The model predicts sequence-level correctness from standardized sequence-level crowding and next-token entropy. Coefficients are estimated using maximum likelihood, and odds ratios (OR) are reported for interpretability. Table~\ref{tab:logit} shows that crowding remains a significant negative predictor of correctness when controlling for entropy, whereas entropy does not exhibit a statistically significant effect. 

\begin{table}[htbp]
\centering
\small
\caption{Logistic regression predicting reasoning correctness. Both predictors are standardized. We report odds ratios (OR) and coefficient estimates.}
\begin{tabular}{lcccc}
\toprule
Predictor & OR & Coef & Std. Err. & $p$-value \\
\midrule
Crowding (z) & 0.29 & $-1.25$ & 0.39 & 0.001 \\
Entropy (z)  & 0.63 & $-0.47$ & 0.41 & 0.26 \\
Intercept    & --   & $-2.39$ & 0.16 & $<0.001$ \\
\bottomrule
\end{tabular}
\label{tab:logit}
\end{table}

\section{Visualizing Crowding and Entropy}
\label{appendix:vce}

\begin{figure}[ht]
  \begin{center}
    \centerline{\includegraphics[width=0.5\columnwidth]{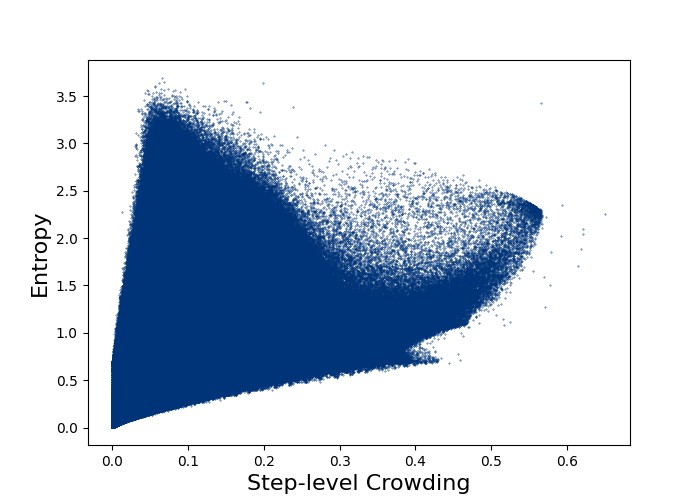}}
    \caption{Scatter plot of step-level embedding-space crowding versus next-token entropy. Each point corresponds to a decoding step. The non-linear and dispersed relationship highlights that crowding and entropy capture distinct properties of the next-token distribution.
    }
    \label{fig:crowding_entropy}
  \end{center}
\end{figure}

Figure~\ref{fig:crowding_entropy} visualizes the relationship between step-level embedding-space crowding and next-token entropy across all decoding steps. While the two quantities are correlated, their relationship is highly non-linear and exhibits substantial dispersion. In particular, similar entropy values can correspond to a wide range of crowding values, and vice versa. This supports the view that crowding captures geometric structure in the next-token distribution that is not fully characterized by entropy alone.

 \section{Derivation of the Correction Strength Factor $\lambda_t$}
\label{appendix:derivation}

% \paragraph{Crowding Correction Target.}
To control the overall correction strength at each decoding step, we introduce a hyperparameter $\tau \in [0,1]$ that modulates how much total probability mass to suppress from the effective correction set. Specifically, we define the target reduction in mass as
\begin{equation}
\delta_t = \tau \cdot \sum_{i \in S_t} p_{t,i},
\end{equation}
where $p_{t,i}$ denotes the original softmax probability of token $i$. This reflects our design goal: downweighting the $S_t$ region by a fraction $\tau$ of its total original mass, adapting the strength of correction to the  shape of the distribution at each step.

% \paragraph{Correction Factor.}
% Given each token’s crowding score ($S_i = (e^{p_{t,i}} - 1)\mathrm{Crowd}_t^{\text{token}}(i)$), we apply a multiplicative attenuation:

Given each token’s crowding score, we define crowding-aware correction
factor for each token (Eq.~\ref{equ:adjustedstep}), with an intermediate value $C_{t,i}$: 

\begin{equation}
C_{t,i} = (e^{p_{t,i}} - 1) \cdot \mathrm{Crowd}_t^{\text{token}}(i),
\end{equation}
% We then apply a multiplicative attenuation to the token's probability based on this factor.
\begin{equation}
\alpha_{t,i} = \frac{1}{1 + \lambda_t C_{t,i}}.
\end{equation}

Then, the total reduction can be defined as
\begin{equation}
 \Delta_t =\sum_{i \in S_t} p_{t,i} \cdot (1 - \alpha_{t,i}).
\end{equation}

The scaling factor $\lambda_t$
is then determined by solving the following implicit equation, which ensures that the actual reduction $\Delta_t$ matches the target reduction $\delta_t$:

\begin{equation}
\label{equ:equ}
\sum_{i \in S_t} p_{t,i} \cdot \frac{\lambda_t C_{t,i}}{1 + \lambda_t C_{t,i}} = \delta_t.
\end{equation}

%\paragraph{Approximate Closed-Form Solution.}
%Directly solving Equation~\ref{equ:equ} requires numerical optimization. To enable a plug-in implementation, we derive a practical approximation. Let $\mu_t = \mathbb{E}[S_t] = \sum_{i \in S_t} p_{t,i}$. Assuming moderate variance in $S_t$, we approximate the expectation using a mean-field approximation:
% To obtain a closed-form update for efficient, plug-in deployment, we derive a practical approximation. We apply a first-order approximation by replacing each $C_{t,i}$ in the sum with the weighted average $\mu_t = \sum_{i \in S_t} p_{t,i} C_{t,i}$ (or $\mathbb{E}[C_{t,i}]$), leading to:
Direct numerical root-finding for Eq.~\ref{equ:equ} is possible but would undermine the efficiency goal of a plug-in decoder. We therefore derive a practical approximation that admits a closed-form solution. We apply a mean-based approximation by replacing each $C_{t,i}$ in the sum with the weighted average $\mu_t = \sum_{i \in S_t} p_{t,i} C_{t,i}$ (or $\mathbb{E}[C_{t,i}]$), leading to:
\begin{equation}
\sum_{i \in S_t} p_{t,i} \cdot \frac{\lambda_t C_{t,i}}{1 + \lambda_t C_{t,i}}
\approx
\frac{\lambda_t \mu_t}{1 + \lambda_t \mu_t}.
\end{equation}

% Solving for $λ_t$ under this approximation yields the following:
Substituting the approximation into the condition $\Delta_t = \delta_t$ and solving for $\lambda_t$ provides a closed-form update:
\begin{equation}
\frac{\lambda_t \mu_t}{1 + \lambda_t \mu_t} = \delta_t
\quad\Longrightarrow\quad
\lambda_t = \frac{\delta_t}{\mu_t(1 - \delta_t)}.
\end{equation}

% Substituting back $\deita_t = \tau \cdot \sum_{i \in S_t} p_{t,i}$ and $\mu_t = \sum_{i \in S_t} p_{i_t} \cdot C_i$, we obtain the final expression:
Recalling that $\delta_t = \tau \cdot \sum_{i \in S_t} p_{t,i}$ and $\mu_t = \sum_{i \in S_t} p_{t,i} C_{t,i}$, and substituting them into $\lambda_t = \frac{\delta_t}{\mu_t(1 - \delta_t)}$, we arrive at the practical, closed-form solution for the step-level correction strength:

\begin{equation}
\boxed{\lambda_t
=
\frac{\tau \cdot \sum_{i \in S_t} p_i}
{\left(\sum_{i \in S_t} p_{t,i} \cdot C_{t,i}\right)\cdot\bigl(1 - \tau \sum_{i \in S} p_i\bigr)} = \frac{\tau \cdot \sum_{i \in S_t} p_i}
{\mathrm{Crowd}^{\text{step}\dagger}(t) \cdot \bigl(1 - \tau \sum_{i \in S} p_i\bigr)}.}
\end{equation}

%\noindent\textbf{Interpretation.}
%This design gives τ a clear, local probabilistic interpretation: it controls the expected fraction of probability mass removed from region $S_t$. By using this, our method dynamically adapts to the local structure of the distribution at each step, ensuring correction is proportionate to the model's crowding distribution and geometric redundancy.
\paragraph{Interpretation.} This formulation provides $\lambda_t$ and $\tau$ with a clear, local probabilistic interpretation. Together, they determine the expected fraction of probability mass to be removed from the correction set $S_t$. Consequently, our method ensures a specified correction strength with $\tau$ while dynamically adapting the step-wise strength $\lambda_t$ to the distribution structure. This dynamic adaptation guarantees that the degree of intervention is scaled appropriately to the prevailing crowding and geometric redundancy.

\section{Ablation for Correction Strength Factor}
\label{app:str}
We conducted an ablation study on a subset of AIME25 with 17 samples (Table~\ref{tab:str}). We observed that when $\lambda_t = 10$, the diversity metric improves while the accuracy performance drops significantly. This aligns with our design rationale: without a Correction Strength Factor that dynamically adjusts according to the distribution and crowding, certain decoding steps may be over- or under-adjusted. Even though it brings benefits in diversity, it does so at the cost of accuracy.
\begin{table}[htbp]
\caption{Ablation study on the Correction Strength Factor ($\lambda_t$). Results on a subset of AIME25 (17 samples) show that a fixed $\lambda_t=10$ improves diversity but severely harms accuracy, highlighting the need for dynamic adjustment.
}
    \centering
    \small
    \setlength{\tabcolsep}{0.8mm}
    \begin{tabular}{lcccc}
\toprule 
& \textbf{Avg@32}  & \textbf{Pass@8}  & \textbf{Dist-N} & \textbf{SemDiv} \\

\midrule

% 23.54 & 39.94	22.18 & 35.01

\method{}  &  \textbf{46.32} &	\textbf{65.31} & 48.85 & 16.88   \\
%  35.94 & 57.33 & 50.36  & 16.61 
% \textbf{36.46} & \textbf{58.91} & \textbf{52.03} & \textbf{17.70}
%\cdashline{1-15} 
%\noalign{\vskip 3pt}
\method{} (Fixed $\lambda_t$)  &  43.38	& 61.07 & \textbf{49.73} & \textbf{17.40}    \\
%(25.31-23.75)
%(33.75-31.82)

\bottomrule
\end{tabular}
    \label{tab:str}
\end{table}

\section{Hyperparameter Details}
\label{app:par}

We provide a detailed specification of the hyperparameter $\tau$ used across different models, configuration settings, and benchmarks, as shown in Table~\ref{tab:par}.

\begin{table}[ht]
\centering
\caption{Hyperparameter $\tau$ used across different models, configuration settings, and benchmark.}
\label{tab:model_comparison}
\begin{tabular}{lcccc}
\toprule
\textbf{Model} & \textbf{Temperature/Top-p} & \textbf{Dataset} & \textbf{Hyper} \\
\midrule
\multirow{6}{*}{Qwen3-1.7B} 
& \multirow{3}{*}{temp=1.0, top-p=1.0} & AIME24 & 0.3 \\
& & AIME25 & 0.3 \\
& & HMMT25 & 0.3 \\
\cmidrule(lr){2-4}
& \multirow{3}{*}{temp=0.7, top-p=0.95} & AIME24 & 0.3 \\
& & AIME25 & 0.2 \\
& & HMMT25 & 0.2 \\
\midrule
\multirow{3}{*}{Qwen3-4B} 
& \multirow{3}{*}{temp=1.0, top-p=1.0} & AIME24 & 0.5 \\
& & AIME25 & 0.5 \\
& & HMMT25 & 0.4 \\
\midrule
Hunyuan-1.8B-Instruct 
& temp=1.0, top-p=1.0 & AIME24 & 0.2 \\
\bottomrule
\end{tabular}
\label{tab:par}
\end{table}

%%%%%%%%%%%%%%%%%%%%%%%%%%%%%%%%%%%%%%%%%%%%%%%%%%%%%%%%%%%%%%%%%%%%%%%%%%%%%%%
%%%%%%%%%%%%%%%%%%%%%%%%%%%%%%%%%%%%%%%%%%%%%%%%%%%%%%%%%%%%%%%%%%%%%%%%%%%%%%%

\end{document}